\documentclass{svjour3}
\usepackage{graphicx}
\usepackage{amssymb}
\usepackage{amsmath}
\usepackage{natbib}
\usepackage{hyperref}
\usepackage{booktabs}
\usepackage{chemformula}
\usepackage{bbm}
\usepackage{multirow}
\usepackage{booktabs} 
\usepackage{xspace}
\usepackage{paralist}
\usepackage[normalem]{ulem}
\usepackage{array}
\usepackage{color}
\usepackage[noend]{algpseudocode}
\usepackage[ruled,vlined,linesnumbered,noend]{algorithm2e}
\usepackage[utf8]{inputenc}
\usepackage{mathabx}
\usepackage{hhline,xcolor}
\usepackage{colortbl}

\usepackage{stmaryrd}

\newcommand{\graphs}{$\mathcal{G}$}

\newcommand{\supp}{{\bf Supp}}
\newcommand{\activate}{{\bf Activate }}
\newcommand{\suppmin}{{\rm minSI }}

\newcommand{\method}{\textsc{\scriptsize INSIDE-GNN}}
\newcommand{\algo}{\textsc{\scriptsize INSIDE-SI}}
\newcommand{\algob}{\textsc{\scriptsize INSIDE-GNN}}
\newcommand{\modif}[1]{{#1}}

\titlerunning{On GNN explanability with activation rules}
\title{On GNN explanability with activation rules}
\date{}
\author{Luca Veyrin-Forrer$^{1}$, Ataollah Kamal$^{1}$, Stefan Duffner$^{1}$, Marc Plantevit$^{2}$ and Céline Robardet$^{1}$}
\authorrunning{Luca Veyrin-Forrer et al.}
\institute{$^{1}$ \modif{Univ Lyon, INSA Lyon, CNRS, UCBL, LIRIS, UMR5205, F-69621 Villeurbanne,
France}\\
  $^{2}$ Laboratoire de Recherche  de l’EPITA (LRE), Le Kremlin-Bic\^etre, 94276, France\\
}
\begin{document}

\maketitle
\begin{abstract}
GNNs are powerful models based on node representation learning that
perform particularly well in many machine learning problems related to graphs.
The major obstacle to the deployment of GNNs is mostly a
problem of societal acceptability and trustworthiness, properties
which require making explicit the internal functioning of such
models. Here, we propose to mine activation \modif{rule}s in the hidden
layers to understand how the GNNs perceive the world. The problem is
not to discover activation \modif{rule}s that are individually highly
discriminating for an output of the model.
Instead, the challenge is to provide a small set of \modif{rule}s
that cover all input graphs. To this end, we introduce the subjective
activation pattern domain. We define an effective and principled
algorithm to enumerate activations \modif{rules} in each hidden
layer. The proposed approach for quantifying the interest of these
\modif{rules} is rooted in information theory and is able to account for
background knowledge on the input graph data.  The activation \modif{rule}s can then be redescribed thanks to pattern languages involving interpretable features.
 We show that the activation \modif{rule}s provide insights on the characteristics used by
the GNN to classify the graphs. Especially, this allows to identify the hidden features built by the GNN through its different layers. Also, these \modif{rule}s can subsequently
be used for explaining GNN decisions.  Experiments on both synthetic
and real-life datasets show highly competitive performance, with up to
$200\%$ improvement in fidelity on explaining graph
classification over the SOTA methods.


\end{abstract}

\section{Introduction}

Graphs are a powerful and widespread data structure used to represent
relational data. One of their specificity is that their underlying
structure is not in a Euclidean space and has not a grid-like
structure \citep{geo17},
characteristics facilitating the direct use of
generic machine learning techniques. Indeed, each node of a graph is
characterized by its features, its neighboring nodes, and recursively
their properties.  Such intrinsically discrete information cannot be
easily used by standard machine learning methods to either predict a
label associated with the graph or a label associated with each node
of the graph.  To overcome this difficulty, Graph Neural Networks
(GNNs) learn embedding vectors \modif{
  to represent
each node $v$ in a metric space and ease comparison between nodes.}
GNN methods \citep{DefferrardBV16,9046288} employ a message propagation strategy that recursively
aggregates information from nodes to neighboring nodes.
%
\modif{This method produces vectors that represent the ego-graphs centered at each node, in such a way that the classification task based on these vectors is optimized. These ego-graphs are induced by nodes that are less than a certain distance from the central node. These distances are equal to the recursion index and correspond to the layer indices in the GNN.}



Although GNNs have achieved outstanding performance in many tasks, a
major drawback is their lack of interpretability.  The last five years
have witnessed a huge growth in the definition of techniques for
explaining deep neural networks
\citep{burkart2021survey,molnar2020interpretable}, particularly for
image and text data. However, the explainability of GNNs has been much
less explored. Two types of approaches have recently been proposed and
have gained certain visibility. Methods based on perturbation \citep{LuoCXYZC020,YingBYZL19}  aim to
learn a mask seen as an explanation of the model decision for a graph
instance. They obtain the best performance for instance explanation. It appears that such masks can lead to unreliable
explanations, and most importantly, can lead to misleading
interpretations for the end-user. One can be tempted to
interpret all the nodes or features of the mask as responsible for the
prediction leading to wrong assumptions. An example of misleading
interpretations is when a node feature is perceived as important for
the GNN prediction, whereas there is no difference between its
distribution within and outside the mask.   XGNN \citep{XGNN} aims at providing model-level explanations by generating a graph pattern that maximizes a GNN output label. Yet,
this method assumes that there is a single pattern for each target which is
not the case in practice when dealing with complex
phenomena. Moreover, these two types of methods query the GNN with perturbed
input graphs to evaluate their impact on the GNN decision and build their
masks from the model output. They do not study the internal mechanisms
of the GNNs, especially the different embedding spaces produced by the
graph convolutions, while we are convinced that the study of GNN
activation vectors may provide new insights on the information used by GNN to achieve the classification of graphs.


In this paper, we consider GNNs for graph \modif{binary} classification.  We
introduce a new method, called \method, that aims at discovering
activation \modif{rules} in each hidden layer of the GNN. An activation
\modif{rule} captures a specific configuration in the embedding space of a
given layer that is considered important in the GNN decision, i.e.,
discriminant for an output label. The problem is therefore not only to
discover highly discriminant activation \modif{rules} but also to provide a
pattern set that covers \modif{all GNN decisions on the input graphs.}  To this end, we define a measure, rooted in the  FORSIED framework \citep{DBLP:conf/kdd/Bie11} to quantify the 
\modif{information provided by a rule relative to that supplied by the rules already extracted.}
The activation pattern set can then support instance-level explanations as
well as providing insights about the hidden features captured and
exploited by the GNN.
\begin{figure}[htb]
\begin{center}
	\includegraphics[width=1\linewidth]{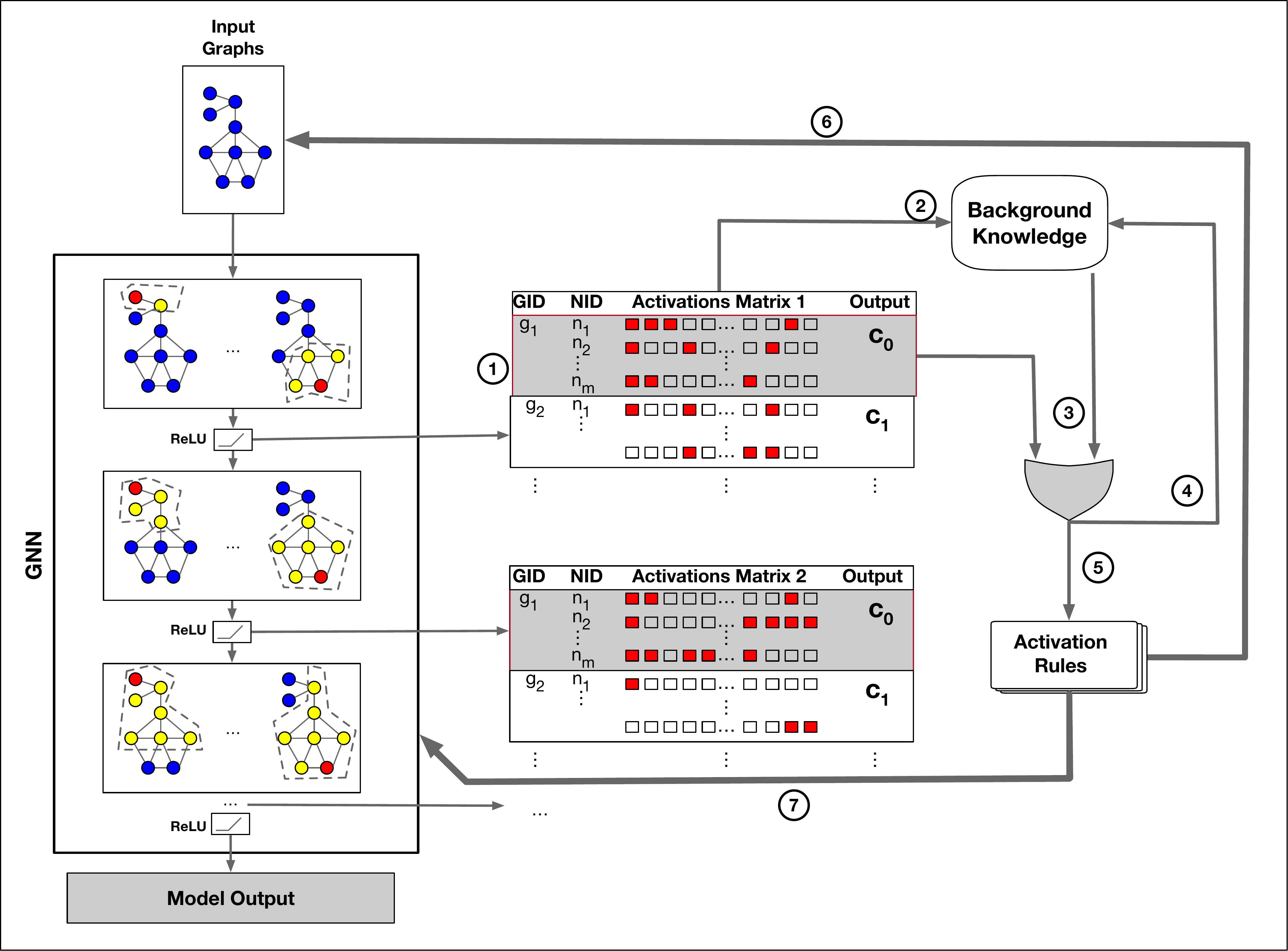}
	\caption{Overview \method: For each layer, (1) \modif{a binary matrix encodes the activation by nodes of embedding vector components. (2) A background model synthesizes the knowledge we have of these data: at the beginning, the probabilities of the components to be activated are independent to the nodes of the graphs. (3) The most informative activation rule (with respect to the background knowledge) is extracted by \method. (4) This rule is integrated into the background knowledge which gradually makes the marginal distributions of the margins of the background model less and less independent.
%
            It is then added to the pattern set (5). Steps (2-5) are repeated until no rule brings significant information about the data in the table. Then, the activation rules are used (6)  to
            support instance level explanations or (7) to provide insights on the model.} }
	\label{fig:overview}
\end{center}
\end{figure}


Fig. \ref{fig:overview} illustrates the main steps of the proposed method. From a trained GNN model and a set of graphs (ideally following the same distribution as the training set), \modif{(1) a binary activation matrix is derived to encode the activation by the graph nodes of the vector components of the GNN. The decision of the GNN is also associated with the nodes. (2) A background model represents the knowledge we have of the matrix data. At the initialization, we have no particular knowledge and we assume that the activations are independent to the nodes of the graphs.
%
  (3) \method{} discovers the most informative  activation rule based on the activation matrix and the background model. (4) The background model is updated to reflect the latest discovered rule that is added to the pattern set (5).
  Steps (2-5) are repeated until no more informative rules are obtained or early termination conditions are reached. (6) The activation pattern set is then used to provide instance-level explanations. To this end, several mask strategies involving nodes that support activation rules are devised. (7) For each activation rule, we use exploratory analysis techniques (e.g., subgroup discovery on graph propositionalization, subgraph mining)  to characterize the nodes supporting the rules and provide interpretable insights on what the GNN really captures. }


Our main contributions are as follows.
After discussing the most important related work in Section \ref{sec:rw} and
	 introducing the novel problem of mining activation rule sets in Section \ref{sec:method}, 
	 we devise a branch-and-bound algorithm that exploits upper-bound-based pruning properties to discover such \modif{rules}.
     We explain how we characterize the activation \modif{rules} with graph properties in Section~\ref{sec:charact}.
	 We report an empirical evaluation in Section \ref{sec:xp} which studies the performance and the potential of the proposed approach for providing instance-level explanations or insights  about the model. \method{} is compared against SOTA explanation methods and outperforms them  by up to $200\%$. We also study the characterization of activation \modif{rules} thanks to interpretable pattern languages. We demonstrate that this allows to obtain good summaries of the hidden features captured by the GNN. Based on this, we eventually compare our approach against a model-level explanation method.




%
%

         \section{Related work}\label{sec:rw}

GNNs are attracting  widespread interest due to their performance in several tasks as node classification, link prediction, and graph classification \citep{wu2020comprehensive}.
Numerous sophisticated techniques allow to improve the performance of such models as graph convolution \citep{KipfW17}, graph attention \citep{VelickovicCCRLB18}, and graph pooling \citep{wang2020second}. However, few researchers have addressed the problem of the GNN explainability compared to image and text domains where a plethora of methods have been proposed \citep{burkart2021survey,molnar2020interpretable}.
As stated in \citep{yuan2020explainability}, existing methods for image classification models explanation cannot be directly applied to not grid-like data: the ones based on the computation of abstract images via back-propagation \citep{SimonyanVZ13} would not provide meaningful results on discrete adjacency matrices; those that learn soft masks to capture important image regions \citep{olah2017feature} will destroy the discreteness property when applied to a graph.



Nevertheless, there have been some attempts to propose
methods for explaining GNNs in the last three years. Given an input graph, the
{\em instance-level} methods aim at providing input-dependent
explanations by identifying the important input features on which the
model builds its prediction. One can identify four different families
of methods. (1) The gradient/feature-based methods -- widely applied
in image and text data -- use the gradients or hidden feature map
values to compute the importance of the input features
\citep{baldassarre2019explainability,PopeKRMH19}. (2) The perturbation-based methods aim at learning a graph mask by investigating the
prediction changes when perturbing the input graphs. GNNExplainer
\citep{YingBYZL19} is the seminal perturbation based method for
GNNs. It learns a soft mask by maximizing the mutual information
between the original prediction and the predictions of the perturbed
graphs.  Similarly, PGExplainer \citep{LuoCXYZC020} uses a generative
probabilistic model to learn succinct underlying structures from the
input graph data as explanations. (3) The surrogate methods explain an input graph by sampling its neighborhood and
learning an interpretable model. GrapheLime \citep{huang2020graphlime}
thus extends the LIME algorithm \citep{ribeiro2016should} to GNN in the
context of node classification. It uses a Hilbert-Schmidt Independence Criterion
Lasso as a surrogate model. However, it does not take into account the graph
structure and  cannot be applied to graph classification models. PGM-Explainer \citep{VuT20} builds a
probabilistic graphical model for explaining node or graph
classification models. Yet, it does not allow to take into
consideration edges in its explanations. These surrogate models can be
misleading because the user tends to generalize beyond its
neighbourhood an explanation related to a local model. Furthermore,
the identification of relevant neighborhood in graphs remains
challenging. Finally, (4) the decomposition-based methods
\citep{PopeKRMH19,abs-2006-03589} start by
decomposing the prediction score to the neurons in the last hidden
layer. Then, they  back-propagate these scores layer by layer until
reaching the input space.  XGNN \citep{XGNN} proposes to provide
a model-Level explanation of GNNs by training a graph generator so that
the generated graph patterns maximize the prediction of the model for
a given label. However, it relies on a strong assumption: each label is
related to only one graph generator which is not realistic when
considering complex phenomena. This is further discussed in Section \ref{sec:xp} based on some empirical evidence.

GNNExplainer, PGExplainer, and PGM-Explainer  are the methods that report the best
performance on many datasets. We will compare our contribution against these methods in
the experimental study. Nevertheless, these methods have some
flaws when used in practice. Discretizing the soft mask (i.e., selecting the most important edges) requires
choosing {a parameter $k$ which is not trivial to set}. Besides,
based on such a mask, the explanation may be misleading because the user
is tempted to interpret what is retained in the mask as responsible
for the decision, and this, even if a node label appears both inside and
outside the mask.

Our method aims to mine some activation patterns in the hidden layers
of GNNs. There exists in the literature some rule extraction methods
for DNNs \citep{tran2018dbnrules}, but not for GNNs. For example,
\citep{tran2018dbnrules} mine association rules from Deep
Belief Networks. Still, their approach suffers from an
explosion of the number of patterns, which makes the results of frequency-based rule mining
mostly unusable in practice. Also, with its focus on DBNs, the method is not directly
applicable to standard GNNs.

\section{\method{} method}
\label{sec:method}
\subsection{Graph Neural Networks \modif{and activation matrix}}
\modif{We consider a set of graphs \graphs{} with labels: $G=(V,E,M)$ with $V$ a set of nodes, $E$ a set of edges in $V\times V$, and $ M$ a mapping between the nodes and the labels, $M\subseteq V\times T$, with $T$ the set of labels.}
The graphs of \graphs{} are classified in two categories \modif{$\{c^0,c^1\}$ by a GNN: GNN$\: :\: $\graphs$\rightarrow\{c^0,c^1\}$}. The GNN takes decisions at the level of each graph on the basis of vectors computed at the level of the nodes.
\modif{These vectors embed nodes into a metric space to ease comparisons.
%
%
%
More precisely, we consider Graph Convolutional Networks (GCN) \citep{KipfW17} with $L$ layers.
GCNs compute vectors $\textbf{h}^\ell_v,\: \ell=1\dots L$ of dimension $K$, an hyperparameter of the method. $\textbf{h}^\ell_v$ represents the ego-graph centered at node $v$ with radius $\ell$. This ego-graph is induced by the nodes that are less than a distance $\ell$ (in number of edges) from $v$. Such vectors are recursively computed by the following formula:}
\begin{eqnarray*}
\textbf{h}^\ell_v = ReLU\left(\textbf{W}_\ell \cdot \sum_{w\in\mathcal{N}(v)}\frac{e_{w,v}}{\sqrt{d_vd_w}} h^{\ell-1}_v\right), \mbox{ with } d_v=\sum_{w\in \mathcal{N}(v)} e_{v,w}. 
\end{eqnarray*}
\modif{$e_{v,w}$ is the weight of the edge between nodes $v$ and $w$, $\mathcal{N}(v)$ is the set of neighboring nodes of $v$ including $v$, ReLU is the rectified linear activation function, and $\textbf{W}_\ell$ are the parameters learned during the model training phase.
  Finally, $\textbf{h}^0_v$ is the initial vector for node $v$ with the one-hot encoding of its labels from set $T$.}
   
\modif{Each vector is of size $K$ and $\ell$ ranges from 0 to $L$ (the maximum number of layers in the GNN), two hyperparameters of the GNN.}
%
%
For a trained GNN, the vectors $\textbf{h}^{\ell}_v$ capture the key characteristics of the corresponding ego-graphs on which the classification is made.
When one of the vector components is of high value, it plays a role in the decision process. More precisely, activated components of the vectors -- those for which $(\textbf{h}^{\ell}_v)_k>0$ -- are combined by the neural network in a path leading to the decision. \modif{We are therefore going to construct the activation matrix corresponding to the activated vector components.}

\begin{definition}[Activation matrix]
  \label{def1}
\modif{The activation matrix $\widehat{H^\ell}$ has dimensions $(n\times K)$, with} $n=\sum_{G_i\in\mathcal{G}} |V_i|$.  $$\widehat{H^\ell}\lbrack v,k\rbrack = \left\lbrace\begin{array}{l} 1\mbox{ if } (\textbf{h}^{\ell}_v)_k>0\\ 0 \mbox{ otherwise}\end{array}\right.$$
\end{definition}

For a given layer $\ell$, the activated components of $\textbf{h}^{\ell}_v$ correspond to the part of the ego-graph centered at $v$ and of radius $\ell$ that triggers the decision. \modif{Therefore, we propose in what follows to identify the sets of components that are activated together for one of the two decisions made by the GNN.}




\begin{figure}[htb]
  \begin{center}
    \begin{tabular}{|c||p{3cm}|c|c||>{\centering}p{1.2em}|>{\centering}p{1.2em}|>{\centering}p{1.2em}|>{\centering}p{1.2em}|>{\centering}p{1.2em}|>{\centering\arraybackslash}p{1.2em}|}
      \hline
&\rotatebox{0}{Graph} & \rotatebox{90}{Output class}& \rotatebox{90}{Node}&\rotatebox{90}{component 1}&\rotatebox{90}{component 2}&\rotatebox{90}{component 3}&\rotatebox{90}{component 4}&\rotatebox{90}{component 5}&\rotatebox{90}{component 6}\\ \hline
      \multirow{5}{5em}{$G_1$} &\multirow{5}{5em}{\includegraphics[scale=0.55]{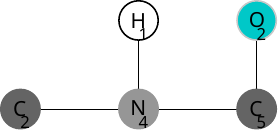}} & \multirow{5}{1em}{$c^1$}&\cellcolor{red}$1$ & \cellcolor[gray]{0.8}$0.1$ & \cellcolor[gray]{0.8}$0.2$ & $0.0$ & $0.0$ & $0.0$ & \cellcolor[gray]{0.8}$0.3$\\
      \hhline{|*{3}{~|}*{7}{-}}
& & & $2$ & $0.0$ & $0.0$ & \cellcolor[gray]{0.8}$0.2$ & \cellcolor[gray]{0.8}$0.2$ & \cellcolor[gray]{0.8}$0.4$ & $0.0$\\ \hhline{|*{3}{~|}*{7}{-}}
& & &\cellcolor{red}$3$ & \cellcolor[gray]{0.8}$0.2$ & \cellcolor[gray]{0.8}$0.1$ & $0.0$ & $0.0$ & $0.0$ & \cellcolor[gray]{0.8}$0.2$\\ \hhline{|*{3}{~|}*{7}{-}}
& & &$4$ & $0.0$ & \cellcolor[gray]{0.8}$0.1$ & $0.0$ & $0.0$ & $0.0$ & \cellcolor[gray]{0.8}$0.2$\\ \hhline{|*{3}{~|}*{7}{-}}
& & &$5$ & $0.0$ & $0.0$ & \cellcolor[gray]{0.8}$0.2$ & $0.0$ & \cellcolor[gray]{0.8}$0.3$ & $0.0$\\\hline
\hline\hline
\multirow{5}{5em}{$G_2$} &\multirow{5}{5em}{\includegraphics[scale=0.55]{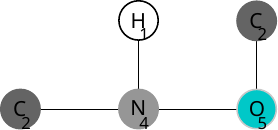}} & \multirow{5}{1em}{$c^1$}&$1$ & \cellcolor[gray]{0.8}$0.1$ & $0.0$ & \cellcolor[gray]{0.8}$0.2$ & \cellcolor[gray]{0.8}$0.2$ & \cellcolor[gray]{0.8}$0.3$ & $0.0$\\
\hhline{|*{3}{~|}*{7}{-}}
& & &\cellcolor{red}$2$ & \cellcolor[gray]{0.8}$0.1$ & $0.0$ & \cellcolor[gray]{0.8}$0.1$ & $0.0$ & \cellcolor[gray]{0.8}$0.1$ & \cellcolor[gray]{0.8}$0.4$\\ \hhline{|*{3}{~|}*{7}{-}}
& & &$3$ & \cellcolor[gray]{0.8}$0.1$ & $0.0$ & \cellcolor[gray]{0.8}$0.2$ & \cellcolor[gray]{0.8}$0.2$ & \cellcolor[gray]{0.8}$0.3$ & $0.0$\\ \hhline{|*{3}{~|}*{7}{-}}
& & &$4$ & $0.0$ & $0.0$ & \cellcolor[gray]{0.8}$0.2$ & $0.0$ & \cellcolor[gray]{0.8}$0.3$ & $0.0$\\ \hhline{|*{3}{~|}*{7}{-}}
& & &$5$ & $0.0$ & $0.0$ & \cellcolor[gray]{0.8}$0.1$ & $0.0$ & \cellcolor[gray]{0.8}$0.1$ & \cellcolor[gray]{0.8}$0.3$\\\hline
\hline\hline
\multirow{5}{5em}{$G_3$} &\multirow{5}{5em}{\includegraphics[scale=0.55]{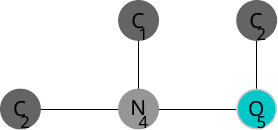}} & \multirow{5}{1em}{$c^0$}&$1$ & $0.0$ & $0.0$ & $0.0$ & $0.0$ & $0.0$ & $0.0$\\ \hhline{|*{3}{~|}*{7}{-}}
& & &$2$ & $0.0$ & \cellcolor[gray]{0.8}$0.3$ & \cellcolor[gray]{0.8}$0.1$ & $0.0$ & \cellcolor[gray]{0.8}$0.1$ & \cellcolor[gray]{0.8}$0.1$\\ \hhline{|*{3}{~|}*{7}{-}}
& & &$3$ & $0.0$ & $0.0$ & $0.0$ & \cellcolor[gray]{0.8}$0.4$ & \cellcolor[gray]{0.8}$0.2$ & $0.0$\\ \hhline{|*{3}{~|}*{7}{-}}
& & &$4$ & $0.0$ & $0.0$ & $0.0$ & $0.0$ & \cellcolor[gray]{0.8}$0.1$ & $0.0$\\
\hhline{|*{3}{~|}*{7}{-}}
& & &$5$ & $0.0$ & \cellcolor[gray]{0.8}$0.2$ & $0.0$ & $0.0$ & \cellcolor[gray]{0.8}$0.1$ & \cellcolor[gray]{0.8}$0.1$\\\hline
\hline\hline
\multirow{5}{5em}{$G_4$} &\multirow{6}{5em}{\includegraphics[scale=0.55]{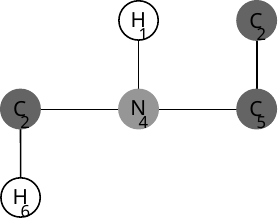}} & \multirow{6}{1em}{$c^0$}&$1$ & $0.0$ & \cellcolor[gray]{0.8}$0.2$ & $0.0$ & $0.0$ & $0.0$ & \cellcolor[gray]{0.8}$0.3$\\
\hhline{|*{3}{~|}*{7}{-}}
& & &$2$ & $0.0$ & $0.0$ & \cellcolor[gray]{0.8}$0.2$ & \cellcolor[gray]{0.8}$0.2$ & \cellcolor[gray]{0.8}$0.3$ & $0.0$\\
\hhline{|*{3}{~|}*{7}{-}}
& & &$3$ & $0.0$ & \cellcolor[gray]{0.8}$0.1$ & $0.0$ & $0.0$ & $0.0$ & \cellcolor[gray]{0.8}$0.1$\\ \hhline{|*{3}{~|}*{7}{-}}
& & &$4$ & $0.0$ & \cellcolor[gray]{0.8}$0.1$ & $0.0$ & $0.0$ & $0.0$ & \cellcolor[gray]{0.8}$0.1$\\ \hhline{|*{3}{~|}*{7}{-}}
& & &$5$ & $0.0$ & $0.0$ & \cellcolor[gray]{0.8}$0.1$ & $0.0$ & \cellcolor[gray]{0.8}$0.2$ & $0.0$\\ \hhline{|*{3}{~|}*{7}{-}}
& & &$6$ & $0.0$ & \cellcolor[gray]{0.8}$0.2$ & $0.0$ & $0.0$ & $0.0$ & $0.0$\\\hline
\hline\hline
\end{tabular}
  \end{center}
  \caption{\label{fig:ex} Toy example: The internal GNN representation of 4 graphs on the third layer with $K=6$. }
\end{figure}

\modif{\subsection{Activation rules discovery}}

We propose to adopt a subgroup discovery approach to identify sets of vector components that are mostly activated in the graphs having the same GNN decision.


\begin{definition}[Activation rule and support]
  \label{def2}
 \modif{An activation rule $A^\ell\rightarrow c$ is composed of a binary vector $A^\ell$ of size $K$ and $c\in\{c^0, c^1\}$ a decision class of the GNN. A graph $G_i=(V_i,E_i, L_i)\in\mathcal{G}$ activates the rule if there is a node $v$ in $V_i$ such that $\widehat{H^\ell}[v,k]=(A^\ell)_k$, for $(A^\ell)_k=1$ and $ k\in\llbracket 1,K\rrbracket$. It is denoted $\activate(A^\ell\rightarrow c,v)$.}
  
    The activated graphs with GNN decision $c$
  form the support of the rule:
  \begin{eqnarray*}
    \supp(A^\ell\rightarrow c,\mathcal{G})&=&
    \lbrace
    G_i\in \mathcal{G}   \mid\exists v\in V_i,\: \activate(A^\ell\rightarrow c,v)
    \mbox{ and } GNN(G_i)=c
    \rbrace 
  \end{eqnarray*}

\end{definition}

\begin{example}
Fig.~\ref{fig:ex} presents the internal GNN representation $(\bold{h}^3_v)$ of 4 graphs on the third layer, where $K=6$. Non-null components (grey cells) are considered as activated and encoded by a '1' value in the binary activation matrix $\widehat{H^3}$. The pattern $A^3=(1,0,0,0,0,1)$ is activated by nodes 1 and 3 of $G_1$ and node 2 of $G_2$. Thus, $\activate(A^3\rightarrow 1,v_1)=True$, $\activate(A^3\rightarrow 1,v_3)=True$ for $G_1$ and $\activate(A^3\rightarrow 1,v_2)=True$ for $G_2$. $\supp(A^3\rightarrow 1,\mathcal{G})=\{G_1,G_2\}$.
\end{example}

Hence, activated \modif{rules} are more interesting if their supports are largely
homogeneous in term of GNN decisions, i.e. the graphs of the support are
mainly classified either in class $c^0$ or in class $c^1$.
We propose to measure the interestingness of these rules in a subjective manner. It makes possible to take into account {\it a priori} knowledge on activation components, but also to perform an iterative extraction of the rules and thus limiting the redundancy between them. These notions are explained below.

\modif{\subsubsection{Measuring the interest of an activation rule}}

\modif{The question now is how to evaluate the interest of the activation rules so as to obtain a set of non-redundant rules. One way to achieve this is to model the knowledge extracted from the activation matrix into a background model and to evaluate the interest of a rule by the knowledge it brings in relation to it. This is what the FORSIED framework \citep{DBLP:conf/kdd/Bie11} does. It proposes an operational way to define the background model and to evaluate the subjective}\footnote{\modif{Subjective means relative to the background knowledge model.}} \modif{interest of a pattern by using information theory to quantify both its informativeness and complexity.}

\modif{We consider the
  discrete random variable $H^\ell\lbrack v,k\rbrack$ associated to the activation matrix  $\widehat{H^\ell}\lbrack v,k\rbrack$\footnote{\modif{We use hats to signify the empirical values.}}, and we model the background knowledge by the probability $P(H^\ell\lbrack v,k\rbrack=1)$.}
Intuitively, the information content (IC) of an activation \modif{rule} should increase when
its components are unusually
activated for the nodes in the graphs of its support
(it is unlikely that these components are activated when considering a random node, while this probability increases when considering graphs supporting the pattern).

\modif{Thus, given the probabilities
$P(H^\ell\lbrack v,k\rbrack=1)$ and with the assumption that all $H^\ell\lbrack v,k\rbrack$ are independent of each other, 
we can
evaluate
the interest of a rule by the product of $P(H^\ell\lbrack v,k\rbrack=1)$ for $v$ activated by the rule and $k$ such that $(A^\ell)_k=1$. Equivalently, we use the negative log-probability.
The more probable the pattern -- and therefore the less interesting -- the smaller this value. As there may exist several nodes activated in a single graph, we choose the one maximizing the measure.
}
\modif{This is formalized in the definition below.}


\begin{definition}[Rule information content]
  \modif{Given a probabilistic background model $P$, the information provided by a rule $A^\ell\rightarrow c$ to characterize a set of graphs \graphs{} is measured by}
  $$IC(A^\ell\rightarrow c,\mathcal{G})=\sum_{G_i\in \supp (A^\ell\rightarrow c,\mathcal{G})}\max_{v\in V_i}-\sum_{k\: s.t.\: (A^\ell)_k=1}\log(P(H^\ell\lbrack v,k\rbrack=1))$$
\end{definition}
\begin{example}
\modif{Considering the example of Fig.~\ref{fig:ex} and the rule $A^3\rightarrow 1$ with $A^3=(1, 0, 0, 0, 0, 1)$ and the probabilistic background model $P$ given in Table~\ref{tab:exBG},  $IC(A^3\rightarrow 1,\mathcal{G})=-\log(0.72)-\log(0.34)-\log(0.99)-\log(0.47)=3.13$.}
\end{example}

A pattern with a large IC is more informative, but it may be more difficult for the user to assimilate it, especially when its description is complex. To avoid this drawback,
the IC value is contrasted by its
description length which measures the complexity of communicating the
pattern to the user. The higher the number of components in $A^\ell$, the more
difficult to communicate it to the user.

\begin{definition}[Description length of a rule]
  \modif{The description length of a rule is evaluated by
  }
$$DL(A^\ell\rightarrow c)=\alpha(\vert A^\ell\vert)+\beta$$ 
 \modif{ with $|A^\ell|$ the L1 norm of $A^\ell$}, $\alpha$ the cost for the user to assimilate each component and $\beta$ a fixed cost for the pattern. We set $\beta= 1$ and $\alpha= 0.6$, as the constant parameter $\beta$ does not influence the relative ranking of the patterns, and with a value of 1, it ensures that the DL value is greater than 1. With $\alpha=0.6$, we express a slight preference toward shorter patterns.
\end{definition}

\begin{example}
  \modif{
    $DL(A^3\rightarrow 1)=2\alpha+\beta=2.2$.}
\end{example}

%
%
%
%

\modif{The subjective interestingness measure is
  defined as the trade-off between IC and DL.}

%


\begin{definition}[Subjective interestingness of a rule]
  The subjective interestingness of a rule on the whole set of graphs \graphs{} is defined by
  $$SI(A^\ell\rightarrow c,\mathcal{G})=\frac{IC(A^\ell\rightarrow c,\mathcal{G})}{DL(A^\ell\rightarrow c)}$$
\end{definition}
\modif{However, in order to identify rules specific to a GNN decision, we consider the difference of subjective interestingness of the measure evaluated on the two groups of graphs.}

\begin{definition}[Differential measure of subjective interest]
  If we denote by $\mathcal{G}^0$ (resp. $\mathcal{G}^1$) the graphs $G_i\in \mathcal{G}$ such that GNN$(G_i)=c^0$ (resp. GNN$(G_i)=c^1$), the subjective interest of the rule $A^\ell\rightarrow c$ with respect to the classes is evaluated by
\modif{$$SI\_SG(A^\ell\rightarrow c)=\omega_c \: SI(A^\ell\rightarrow c,\mathcal{G}^c)-\omega_{1-c}\:  SI(A^\ell\rightarrow c,\mathcal{G}^{1-c}).$$}
The weights $\omega_0$ and $\omega_1$ are used to counterbalance the measure in unbalanced decision problems. The rational is to reduce the SI values of the majority class.
%
We set $\omega_0=\max(1,\frac{\vert \mathcal{G}^1\vert}{\vert \mathcal{G}^0\vert})$ and $\omega_1=\max(1,\frac{\vert \mathcal{G}^0\vert}{\vert \mathcal{G}^1\vert})$.
\end{definition}

\begin{example}
  \modif{
    $SI\_SG(A^3\rightarrow 1)=SI(A^3\rightarrow 1,\mathcal{G}^{1})- SI(A^3\rightarrow 1,\mathcal{G}^{0})=\frac{IC(A^3\rightarrow 1,\mathcal{G}^1)}{DL(A^3\rightarrow 1)}-\frac{IC(A^3\rightarrow 1,\mathcal{G}^0)}{DL(A^3\rightarrow 1)}=\frac{3.13}{2.2}-0$.}
\end{example}

\subsubsection{\modif{Computing the background model}}
\modif{The background model is initialized with basic assumptions about the activation matrix and updated as rules are extracted.}

\begin{definition}[Initial background model]
  \label{def6}
  \modif{Some components can be activated more than others on all the graphs, or some nodes can activate a variable number of components. We assume that this information is known and use it to constrain the initial background distribution $P$:}
\begin{eqnarray*}
  \sum_{v} P(H^\ell\lbrack v,k\rbrack=1)&=& \sum_{v} P(\widehat{H^\ell}\lbrack v,k\rbrack=1),\\ 
  \sum_{k} P(H^\ell\lbrack v,k\rbrack=1)&=& \sum_{k} P(\widehat{H^\ell}\lbrack v,k\rbrack=1).
\end{eqnarray*}
However, these constraints do not completely
specify the probability matrix.
Among all the probability distributions satisfying these constraints, we choose the one with the maximum entropy. Indeed, any
distribution $P$ with an entropy lower than the maximum entropy
distribution effectively injects additional knowledge, reducing
uncertainty unduly. The explicit mathematical MaxEnt model solution can be found in \citep{debie09}.
\end{definition}

\modif{The corresponding initial background model of example of Fig.~\ref{fig:ex} is given in Table~\ref{tab:exBG}.}
\begin{table}[htb]
  \begin{center}
  \begin{tabular}{|c|c||c|c|c|c|c|c|}
    \hline
    \rotatebox{0}{Graph} &  \rotatebox{90}{Node}&\rotatebox{90}{component 1}&\rotatebox{90}{component 2}&\rotatebox{90}{component 3}&\rotatebox{90}{component 4}&\rotatebox{90}{component 5}&\rotatebox{90}{component 6}\\ \hline
    \hline
    \multirow{5}{5em}{$G_1$}
    & 1&0.729 &0.556&0.556&0.507&0.346&0.346 \\
 &2&0.729&0.556&0.556&0.507&0.346&0.346 \\
& 3&0.729&0.556&0.556&0.507&0.346& 0.346 \\
 &4&0.527&0.402&0.402& 0.366&0.250&0.250\\
 &5&0.527&0.402& 0.402& 0.366&0.250&0.250\\
 \hline
\multirow{5}{5em}{$G_2$} & 1&0.999&0.762&0.762&0.695&0.474&0.474\\
 &2&0.999&0.762&0.762&0.695&0.474&0.474\\
 &3&0.999&0.762&0.762&0.695&0.474&0.474\\
 &4&0.527&0.402& 0.402& 0.366&0.250&0.251\\
 &5&0.729&0.556&0.556&0.507&0.346& 0.346 \\
 \hline
 \multirow{5}{5em}{$G_3$} &
 1&0.256&0.195&0.195&0.178&0.122&0.122\\
 &2&0.999&0.762&0.762&0.695&0.474&0.474\\
 &3&0.527&0.402& 0.402& 0.366&0.250&0.250\\
 &4&0.374&0.285&0.285&0.259&0.177&0.177\\
 &5&0.730&0.556&0.556&0.507&0.346 & 0.346 \\
 \hline
 \multirow{6}{5em}{$G_4$} &
 1&0.527&0.402& 0.402& 0.366&0.250&0.250\\
& 2&0.729&0.556&0.556&0.507&0.346& 0.346 \\
& 3&0.527&0.402 &0.402& 0.366&0.250&0.250\\
& 4&0.527&0.402 &0.402& 0.366&0.250&0.250\\
& 5&0.527&0.402& 0.402& 0.366&0.250& 0.250\\
& 6&0.374&0.285&0.285&0.259&0.177&0.177\\
   \hline
  \end{tabular}
  \end{center}
  \caption{\label{tab:exBG}\modif{Initial background model $P(H^\ell\lbrack v,k\rbrack=1)$ of example of Fig.~\ref{fig:ex}.}}
\end{table}

\modif{Once a rule $A^\ell\rightarrow c$ has been extracted, it brings some information about the activation matrix that can be integrated into $P$. The model must integrate the knowledge carried by this rule, that is to say that all the components with value 1 of $A^\ell$ are activated by the vertices activating the rule.}

\begin{definition}[Updating the background model]
  \modif{
    The model $P$ integrates the rule $A^\ell\rightarrow c$ as follows:}
  $\forall k$ such that  $(A^\ell)_k=1$ and $v$
  such that $\widehat{H^\ell}[v,k]=(\textbf{A}^\ell)_k$,
  $P(H^\ell\lbrack v,k\rbrack=1)\mbox{ is set to } 1.$
\end{definition}

\subsubsection{Iterative extraction of subjective activation subgroups}

We propose to compute the subjective activation \modif{rules} with an
enumerate-and-rank approach. It consists to compute the \modif{rule}
$A^\ell\rightarrow c$ with the largest $SI\_SG$ value
and to integrate it in the background distribution $P$ to take into
account this newly learnt piece of information.

Algorithm~\ref{algo2}
sketches the method. First, \modif{it sets the output set equal to the
  empty set (line 1) and the $\suppmin$ value to the largest value
  (line 2). $\suppmin$ corresponds to the smallest $SI\_SG$ value of the extracted patterns.
  A stack of size $K$ is initialized line 3. The first
  considered rule $A^\ell$ is initialized as a bit-vector of size $K$
  containing only 0's. It corresponds to the rule with no activated
  components. It has an associate attribute $Pot$ that encodes the
  components that could still be activated for $A^\ell$, as it leads a yet
  unconsidered combination of activated components. Rule $A^\ell$ is then staked in
  $Stack$ (lines 4 to 6).  } Line 7, it computes the background model
$P$ from the activation matrix $\widehat{H^\ell}$ \modif{as defined in
  Definition~\ref{def6}}.  Then, in a loop (lines 8 to 11), it computes
iteratively the rule $A^\ell\rightarrow c$ having the best $SI\_SG(A^\ell\rightarrow c)$ value. Then, this best rule is used
to update the model $P$ (line 11). Indeed, once the \modif{rule} $A^\ell$ is
known, its subjective interest falls down to 0. This consists in
setting the corresponding probabilities to 1.

\begin{algorithm}[htb]
    \caption{\algob($\widehat{H^\ell}$, $c$, $nbPatt$)\label{algo2}}
  \small
  \KwIn{$\widehat{H^\ell}$ the activation matrix (see Definition~\ref{def1}), $c$ the class to be characterized and
    $nbPatt$ the number of patterns to extract.}
 \KwOut{$output$, up to $nbPatt$ best activation rules $A^\ell\rightarrow c$ w.r.t. $SI\_SG$.}

 $output\leftarrow\emptyset$\\
 $\suppmin\leftarrow 0$\\
 $Stack.maxsize \leftarrow K$\\
\modif{ $A^\ell\leftarrow$ a size $K$ bit-vector initialized at 0}\\
 \modif{$A^\ell.Pot\leftarrow$ a size $K$ bit-vector full of 1's}\\
 $Stack[0]\leftarrow A^\ell$\\
 $P\leftarrow Compute\_Model(\widehat{H^\ell})$\\
 \Repeat{(($|output|\geq nbPatt$) \textbf{or} ($\suppmin < 10$))}{
  $A^\ell,\suppmin\leftarrow$ \algo($Stack$, $P$, $c$, \suppmin, 0, $\emptyset$)\\
  $output\leftarrow output\cup A^\ell$\\
  Update\_Model$(P,A^\ell)$\\
  }

\end{algorithm}

\begin{algorithm}[htb]
    \caption{\algo($Stack$, $P$, $c$, $\suppmin$, depth, $Best$)\label{algo1}}
  \small
  \KwIn{ 
    $Stack$ a stack of recursively enumerated patterns at depth $depth$, $P$ the background distribution, $c$ the class to be characterized, 
    $\suppmin$ a dynamic threshold on $SI\_SG(A^\ell\rightarrow c)$.}
 \KwOut{$Best$, the best rule w.r.t. $SI\_SG$.}

 $A\leftarrow Stack[depth]$\\

$Best\leftarrow A$\\

 \If{($(\phi(A) = False)$ or $(UB\_SI(A,P,c) < \suppmin)$}{
   \Return
 }

 \If{($A.Pot = \emptyset$)}{
   \If{$Best =\emptyset$}{
     \If{$SI\_SG(A\rightarrow c) >\suppmin$}{
       $Best\leftarrow A$\\
     }
   }
   \Else{
     \If{($SI\_SG(A\rightarrow c) > SI\_SG(Best\rightarrow c)$)}{
       $Best\leftarrow A$\\
       $\suppmin\leftarrow SI\_SG(Best\rightarrow c)$\\
     }
   }
   }
 \Else{
   $x \leftarrow$ first bit of $A.Pot$ set to 1\\
   $A.Pot[x]\leftarrow 0$\\
   $A[x]\leftarrow 1$\\
   $Stack[depth+1]\leftarrow A$\\
   \algo($Stack$, $P$, $c$, $\suppmin$, depth+1)\\
    $A[x]\leftarrow 0$\\
   $Stack[depth+1]\leftarrow A$\\
   \algo($Stack$, $P$, $c$, $\suppmin$, depth+1)\\
 }
 \Return $Best$, $\suppmin$
\end{algorithm}

Algorithm~\ref{algo1} presents \algo{} that computes the best
\modif{rule with as activated components the one's values of the vector stored in $Stack[depth]$,
  and even more, depending on the recursive process.}  It considers a
pattern $A$ stored in the stack at depth $depth$. $A$ has 5
attributes:
\begin{itemize}
  \item $A.Pot$, a \modif{vector whose one's values represent the activated} components that can be further added to $A$
    during the enumeration process,
    \item $A.G^c$ (resp. $A.G^{1-c}$) the set of
graphs from $\mathcal{G}^c$ (resp. $\mathcal{G}^{1-c}$) that support
$A$,
\item and $A.TG^c$ (resp. $A.TG^{1-c}$) the set of graphs that are
supporting $A$ and all its descendants in the enumeration process (there is a node in these
graphs that activates all the components of $A$ and $A.Pot$).
\end{itemize}
The algorithm computes the closure of $A$ using the function $\phi$. \modif{It consists in
adding activated components to $A$ (set some components of A to 1)}
as long as the set $A.G^c$ of supporting graphs stays
unchanged. \modif{Furthermore, if a component has been removed from $A$ (on
line 14) but can be added later to $A$ (i.e. $\phi(A)\& A.pot\not= A$ with \& the bitwise \textbf{and} operation), $A$ is not closed, the function returns False and the
recursion stops.}

Line 2, a second criterion based on an
upper bound $UB\_SI$ makes the recursion stop if its value is less
that the one of the current best found rule. It relies on the
following property.
\begin{property}
  \modif{Let $A$ and $B$ be two binary vectors of size $K$. The components that are activated for $A$ are also activated for $B$ (i.e., $A\& B= A$, with $\&$ the bitwise \textbf{and} operation). We can upper bound the value $SI\_SG(B\rightarrow c)$ and have}
  $$SI\_SG(B\rightarrow c)\leq UB\_SI(A,P,c)\mbox{ with}$$

  \begin{eqnarray*}
    UB\_SI(A,P,c)&=&
     w_c\frac{\sum_{g\in A.G^c}\max_{v\in V_g}-\sum_{k\: s.t.\: (A\& A.Pot)_k=1}\log(P(H^\ell\lbrack v,k\rbrack=1))}{\alpha(\vert A\vert)+\beta}\\
    &&- w_{1-c}\frac{\sum_{g\in A.TG^{1-c}}\max_{v\in V_g}-\sum_{k\: s.t.\: (A)_k=1}\log(P(H^\ell\lbrack v,k\rbrack=1))}{\alpha(\vert A\& A.Pot\vert)+\beta}\\
  \end{eqnarray*}
  with $|A|$ the L1 norm of $A$.
\end{property}


\begin{proof} To upper bound the measure $SI\_SG(B\rightarrow c)$, we follow the strategy explained in \citep{DBLP:journals/tkdd/CerfBRB09}. Let  $$SI\_SG(B\rightarrow c)=w_c\frac{X}{Y_1}-w_{1-c}\frac{Z}{Y_2}$$ with
\begin{eqnarray*}
  X&=&IC(B,\mathcal{G}^c)=\sum_{G_i\in \supp(B,\mathcal{G}^c)}\max_{v\in V_i}-\sum_{k\: s.t.\: (B)_k=1}\log(P(H^\ell\lbrack v,k\rbrack=1))\\
 Z&=&IC(B,\mathcal{G}^{1-c})=\sum_{G_i\in \supp(B,\mathcal{G}^{1-c})}\max_{v\in V_i}-\sum_{k\: s.t.\: (B)_k=1}\log(P(H^\ell\lbrack v,k\rbrack=1))\\
Y_1&=&Y_2=DL(B)=\alpha(\vert B\vert)+\beta
\end{eqnarray*}
Similarly, we denote the upper bound function by $$UB\_SG(A,P,c)=w_c\frac{\gamma}{\delta}-w_{1-c}\frac{\epsilon}{\eta}.$$

\modif{Therefore, the largest value of $SI\_SG(B\rightarrow c)$ is obtained if:}
\begin{itemize}
\item \modif{$X$ has the maximal possible value, that is to say $B=A\& A.Pot$}
  and all the graphs of $\mathcal{G}^c$ that support $A$, also support $A\& A.Pot$. In that case, we have
  $$\gamma=\sum_{g\in A.G^c}\max_{v\in V_g}- \sum_{k\: s.t.\: (A\& A.Pot)_k=1}\log(P(H^\ell\lbrack v,k\rbrack=1))$$
\item $Y_1$ has the smallest possible value $\alpha(\vert A\vert)+\beta$ (more elements in $B$ will decrease the value of the fraction)
  $$\delta = \alpha(\vert A\vert)+\beta$$
\item $Z$ has the smallest possible value and is computed over $A$, and on the graphs from $\mathcal{G}^c$ that support $A$ and all its descendants ($A.TG^{1-c}$)
  $$\epsilon=\sum_{g\in A.TG^{1-c}}\max_{v\in V_g}-\sum_{k\: s.t.\: (A)_k=1}\log(P(H^\ell\lbrack v,k\rbrack=1))$$
\item $Y_2$ has the value $\alpha(\vert A\& A.Pot\vert)+\beta$ (less elements in $B$ will decrease the value of the function)

  $$\eta=\alpha(\vert A\& A.Pot\vert)+\beta$$
\end{itemize}
It results in the upper bound definition.
\end{proof}

%

\modif{Line 4 of Algorithm~\ref{algo1}, $Best$ is updated as well as $\suppmin$ if there are no more component to enumerate, and if the SI\_SG value of the current rule is better than the one already found. Otherwise (lines 9 to 16), the enumeration continues either 1) by adding a component from $A.Pot$ to $A$ (lines 9-12) and recursively call the function (line 13), or 2) by not adding the component while still removing form $A.Pot$ (line 14) and recursively continue the process (line 16).}




\section{\modif{Characterization of activation rules with subgroups}}
\label{sec:charact}
Once the activation patterns are found, we aim to describe them in an intelligible and accurate way.
We believe that each activation pattern can be linked to hidden features of the graphs, that are captured by the model as being related to the class to be predicted. The objective here is to make these features explicit. For this, we seek to characterize the nodes that support the activation pattern, and more precisely to describe the singular elements of their neighborhoods. 
%
Many pattern domains can be used to that end. In the following, we consider two of them: one based on numerical descriptions and the other one based on common subgraphs. In order to characterize the subgraphs centered on the nodes of the activation pattern support (called ego-graphs) in a discriminating way compared to the other subgraphs, we extend the well-known GSPAN algorithm \cite{yan2002gspan} so that it takes into account subgroup discovery quality measure.  



\subsection{ Numerical subgroups}
In this approach, we propose to describe each node that supports a
given activation pattern by some topological properties\footnote{These
  attributes are computed with Networkx Python Library
  \url{https://networkx.org/}.}. We choose to consider its degree, its betweenness
centrality value, its clustering-coefficient measure, and the number of
triangles it is involved in, as characteristic features. These properties can be extended to the
whole \modif{ego-graph} by aggregating the values of the neighbors. We
consider two aggregation functions: the sum and the mean. Thanks to
these properties, we make a propositionalization of the nodes of the
graphs and we consider as target value the fact that the node belongs to
the support of the activation \modif{rule} (labeled as a positive example) or not (labeled as a negative example).
\modif{Therefore, we have a matrix $\mathcal{D}$ whose rows denote graph nodes and columns correspond to numerical attrbutes describing the position of the node in its graph: $\mathcal{D}\lbrack v\rbrack\in\mathbb{R}^p$, with $p$ the number of attributes. $\mathcal{D}$ is split in two parts $\mathcal{D}^0$ and $\mathcal{D}^1$ with $v\in \mathcal{D}^c$ iff
$\activate(\textbf{A}^\ell\rightarrow c,v)$.}

To identify the specific descriptions of the support nodes, we propose to use a subgroup discovery method in numerical data.
%
%
It makes it possible to find restrictions on numerical attributes (less or greater than a numerical value) that characterize the presence of a node within the support of the activation \modif{rule. A numerical pattern has the form $\bigtimes_{i=1}^p \lbrack a_i,b_i\rbrack$ (i.e. the pattern language) and a graph node supports the pattern if $\forall i=1\dots p,\: a_i\leq \mathcal{D}\lbrack v,i\rbrack\leq b_i$.}

To discover such subgroups, we use the pysubgroup library \cite{DBLP:conf/pkdd/Lemmerich018}.

\subsection{Graph subgroups}
Another approach consists to characterize activation \modif{rules} by subgraphs that are common among
positive examples in contrast to the negative ones. 
 To this end, we consider as positive examples the \modif{ego-graphs} (with a radius equal to the layer) of nodes that support the activation pattern of interest. By taking the radius into account, we are not going beyond what the model can actually capture at this layer. The negative examples are the graphs in \graphs{} for which none of their vertices support the activation pattern. Hence, \modif{$\mathcal{D}$ is a set of graph nodes $v$ associated to ego-graphs $\mathcal{E}_g=(V_g,E_g,L_g)$. $\mathcal{D}$ is split into $\mathcal{D}^0$ and $\mathcal{D}^1$ with $v\in \mathcal{D}^c$ iff $\activate(\textbf{A}^\ell\rightarrow c,v)$.}

 \modif{A graph pattern has the form $G=(V,E,L)$ (i.e. the pattern language) and a graph node supports the pattern if there exists a graph isomophism between $\mathcal{E}_g=(V_g,E_g,L_g)$ and its ego-graph $\mathcal{E}_g=(V_g,E_g,L_g)$.}

\subsection{Quality measure and algorithms}
As for the identification of activation patterns, we could have used
subjective interestingness measure to characterize the supporting
ego-graphs of the activation patterns. However, we opt for a more
usual measure, the Weighted Relative Accuracy \citep{WRACC}. \modif{Given a pattern $P$ of a given language, a dataset  $\mathcal{D}$ split into $\mathcal{D}^0$ and $\mathcal{D}^1$ and a $\supp(P,\mathcal{D})$ measure that gives all the graph nodes supporting the pattern $P$ in the data $\mathcal{D}$, the WRAcc measure} 
%
%
$$WRAcc(P,c)= \frac{|\supp(P,\mathcal{D})|}{|\mathcal{D}|} \left(\frac{|\supp(P,\mathcal{D}^c)|}{|\supp(P,\mathcal{D})|} - \frac{|\mathcal{D}^c|}{|\mathcal{D}|}\right)$$
\modif{gives high values to patterns that are mainly supported by nodes of $\mathcal{D}^c$ compared to the whole dataset $\mathcal{D}$.
  Then, we use off-the-shelf algorithms to discover the best subgroups. We compute patterns $P$ such that
  \begin{eqnarray}
    \label{eq1}
    WRAcc(P,c)\geq \min\_{WRAcc}\mbox{ and }|\supp(P,\mathcal{D})|\geq\min\_sup
  \end{eqnarray}
or just the subgroup with the highest WRAcc value.}

For the numerical subgroups, we use \texttt{Pysubgroup} library \citep{DBLP:conf/pkdd/Lemmerich018}.
For graph subgroup dicovery, we integrate the WRAcc measure
into the GSPAN algorithm \citep{yan2002gspan}.
As $WRAcc$ measure is not anti-monotone, we use the following upper-bound instead of the $WRAcc$ for pruning:
$$
UB(P,c) = \frac{|\supp(P, \mathcal{D})|}{|\mathcal{D}|}\left( 1 - \frac{\max\left(min\_sup, |\mathcal{D}^c|\right)}{|\mathcal{D}|} \right)
$$ 
If $min\_sup < |\mathcal{D}^c|$, then we have $UB(P,c)=\frac{|\supp(P, \mathcal{D})|}{|\mathcal{D}|}\left( 1 - \frac{|\mathcal{D}^c|}{|\mathcal{D}|} \right)$. Since $\frac{|\supp(P, \mathcal{D}^c)|}{|\supp(P, \mathcal{D})|} \leq 1$, $WRAcc(P,c) \leq UB(P,c)$. In the other case, we have:
\begin{align*}
&\frac{|\supp(P, \mathcal{D}^c)|}{|\supp(P, \mathcal{D})|} - \frac{|\mathcal{D}^c|}{\mathcal{D}} \leq \frac{|\supp(P, \mathcal{D})|}{|\supp(P, \mathcal{D})|} - \frac{min\_sup}{|\mathcal{D}|}
\Leftrightarrow\\
&\frac{min\_sup}{|\mathcal{D}|} - \frac{|\mathcal{D}^c|}{|\mathcal{D}|} \leq \frac{|\supp(P, \mathcal{D})|}{|\supp(P, \mathcal{D})|} - \frac{|\supp(P, \mathcal{D}^c)|}{|\supp(P,\mathcal{D})|}
\Leftrightarrow\\
&\frac{1}{|\mathcal{D}|}(min\_sup - |\mathcal{D}^c|) \leq \frac{1}{|\supp(P,\mathcal{D})|}(|\supp(P,\mathcal{D})|- |\supp(P, \mathcal{D}^c)|)
\end{align*}
The last inequality holds since $\frac{1}{|\mathcal{D}|} \leq \frac{1}{|\supp(P,\mathcal{D})|}$, $\min\_sup \leq |\supp(P,\mathcal{D})|$, and finally $|\mathcal{D}^c| \geq |\supp(P,\mathcal{D}^c)|$.

Since $UB$ is not dependent to the $\supp(P, \mathcal{D}^c)$, when $|\mathcal{D}^c|$ is much lower than the $|\mathcal{D}|$, this upper bound is not tight. We can use another upper bound which is dependent to the $|\mathcal{D}^c|$. Let us call this upper bound $UB2$:
$$
UB2(P,c) = \frac{|\supp(P, \mathcal{D}^c)|}{|\mathcal{D}|} - \frac{min\_sup}{|\mathcal{D}|}\times \frac{|\mathcal{D}^c|}{|\mathcal{D}|}
$$
Since except $\supp(P, \mathcal{D}^c)$ everything is constant, and $\supp(P,\mathcal{D}^c)$ is anti-monotone, $UB2$ is anti-monotone too. To show that $UB2$ is an upper bound for $WRAcc$, note that $\frac{min\_sup}{|\mathcal{D}|}\times \frac{|\mathcal{D}^c|}{|\mathcal{D}|} \leq \frac{|\supp(P, \mathcal{D})|}{|\mathcal{D}|}\times \frac{|\mathcal{D}^c|}{|\mathcal{D}|}$ and the first terms of $WRAcc$ and $UB2$ are equal.
In our algorithm we use $UB3(P,c)=\min\lbrace UB2(P,c), UB(P,c)\rbrace$ as  upper bound for the $WRAcc$.

\section{Experimental study}\label{sec:xp}

In this section, we evaluate \method{} through several experiments. We
first describe synthetic and real-world datasets and the experimental
setup. Then we present a quantitative study of the patterns provided
by \method. Next, we show the experimental results on explanations of
graph classification against several SOTA methods. Finally, we report results on the characterization of activation \modif{rules} by human understandable descriptions of what GNN models capture.
\method{} has been implemented in
Python and the experiments have been performed on a machine equipped
with 8 Intel(R) Xeon(R) W-2125 CPU @ 4.00GHz cores 126GB main memory,
running Debian GNU/Linux.
\href{https://www.dropbox.com/sh/jsri7jbhmkw6c8h/AACKHwcM3GmaPC8iBPMiFehCa?dl=0}{The code and the data are available}\footnote{\url{https://www.dropbox.com/sh/jsri7jbhmkw6c8h/AACKHwcM3GmaPC8iBPMiFehCa?dl=0}}.

\subsection{Datasets and experimental setup }
Experiments are performed on six  graph classification datasets  whose main characteristics are given in Table \ref{tab:data}.
BA2 \citep{YingBYZL19} is a synthetic dataset generated with Barabasi-Albert graphs and hiding either a 5-cycle (negative class) or a ``house'' motif (positive class).  The other datasets (Aids \citep{DBLP:journals/corr/abs-2007-08663}, BBBP\citep{DBLP:journals/corr/WuRFGGPLP17}, Mutagen \citep{DBLP:journals/corr/abs-2007-08663}, DD \citep{DandD}, Proteins \citep{DBLP:conf/ismb/BorgwardtOSVSK05})
depict real molecules and the class identifies  important properties in Chemistry or Drug Discovery (i.e., possible activity against HIV, permeability and  mutagenicity).

A  3-convolutional layer GNN (with $K=20$) is trained on each dataset \modif{using 80\% of the data (train set). The hyperparameters are chosen using a grid-search on other 10\% of the data (validation set). The learned GNN are tested on the remaining 10\% of the data (test set). The corresponding accuracy values are reported in Table \ref{tab:data}.}
\method{} mines the corresponding GNN activation matrices to discover subjective activation pattern set. We extracted at most ten patterns per layer and for each output value, with a $SI\_SG$ value greater than 10.

\begin{table}[htb]
  \begin{center}
\caption{Main characteristics of the datasets.}
\label{tab:data}
\scalebox{1}{
\begin{tabular}{|l||c|l|m{0.75cm}|m{0.75cm}|m{0.75cm}|m{0.75cm}|m{0.75cm}|}
\hline
Dataset & \#Graphs & (\#neg,\#pos)	& Avg. Nodes	&Avg. Edges & Acc. (train) &Acc. (test) &Acc. (val) \\ \hline\hline
BA2(syn)
& 1000 &  (500, 500) & 25 & 50.92 & 0.995 & 0.97 & 1.0\\
Aids
& 2000 & (400, 1600) & 15.69 & 32.39 & 0.989 & 0.99 & 0.975\\
BBBP& 1640 & (389, 1251) & 24.08 & 51.96 & 0.855 & 0.787 & 0.848\\
Mutagen  & 4337 & (2401, 1936) & 30.32 & 61.54 & 0.815 & 0.786 & 0.804\\
DD & 1168 & (681, 487) & 268 & 1352 & 0.932 & 0.692 & 0.760\\
Proteins & 1113 & (663, 450) & 39 & 145 & 0.754 & 0.768 & 0.784\\ \hline
\end{tabular}
}
\end{center}
\end{table}

\subsection{Quantitative study of activation \modif{rules}}
Table \ref{tab:mim} reports general indicators about the discovery of
activation \modif{rules} by \method. The execution time ranges from few
minutes for simple task (i.e., synthetic graphs) to two days for more
complex ones (i.e., DD). It shows the feasibility of the proposed
method. Notice that this process is performed only once for each
model.

\modif{To assess whether the set of extracted rules represent the GNN
  well and in its entirety, we used the rules to describe the input
  graphs (i.e. the graphs (in row) are described by the rules (in columns) and the
  data matrix contains the number of graph nodes supporting the
  corresponding rule). We then learned the simple and
  interpretable model that is the decision tree. Thus, from only the
  knowledge of the number of nodes of a graph supporting each of the
  rules, we can see, in Table \ref{tab:mim} last column, that the decision tree can
  mimic the GNN decision output with high accuracy.}
Obviously, we do not provide an interpretable model
yet, since the decision tree is based on the patterns that capture
sets of activated components of the GNN. Nevertheless, the results demonstrate
that the pattern set returned by \method{} captures the inner workings
of GNNs well.

\begin{table}[htb]
  \begin{center}
\caption{Execution time, number of discovered patterns by \method{}
  and the ability of the pattern set to mimic GNN\modif{: the accuracy of a decision tree with activation rules as features and measured on a test set.
    The class variable is the GNN
    output $y_i$. The closer  $Acc(DT^P,y_i)$ to $1$, the better the
  mimicry.}} \label{tab:mim} \scalebox{1}{
\begin{tabular}{|l||l|l|l|}
	\hline
Dataset & Time (s) & \# Act. Rules & $Acc(DT^P,y_i)$ \\ \hline\hline
BA2(syn) & 180 & 20 &0.98 \\
Aids & 5160 & 60 &0.96 \\
BBBP & 6000 & 60 & 0.89\\
Mutagen & 41940 & 60 &0.87 \\
DD & 212400  & 47 & 0.86\\
Proteins & 8220  & 29 & 0.87  \\
	\hline
	\end{tabular}
        }
\end{center}
\end{table}

\begin{figure}[htb]
\begin{center}
\begin{tabular}{ccc}
 \includegraphics[width=.28\linewidth]{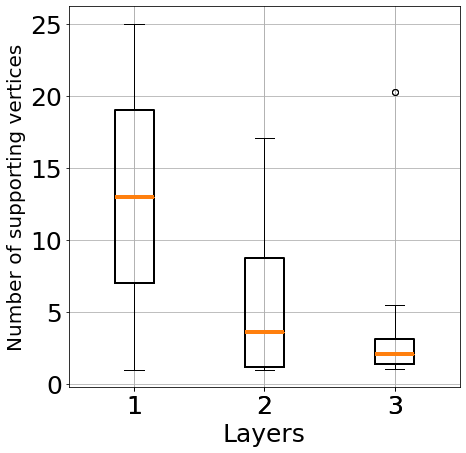}&
\includegraphics[width=.3\linewidth]{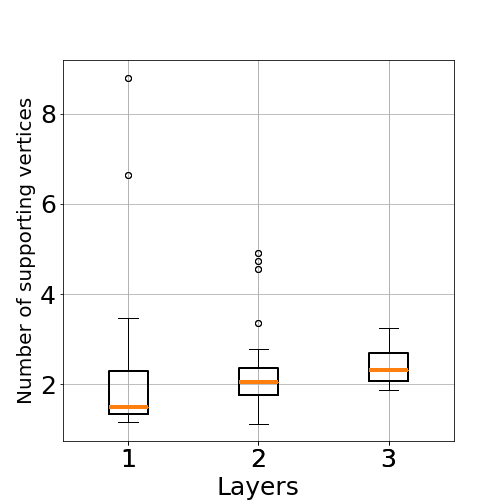}&
\includegraphics[width=.3\linewidth]{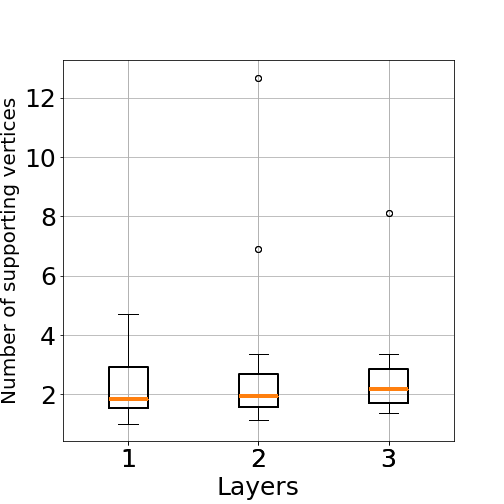}\\
BA2&  Aids & BBBP \\
\includegraphics[width=.3\linewidth]{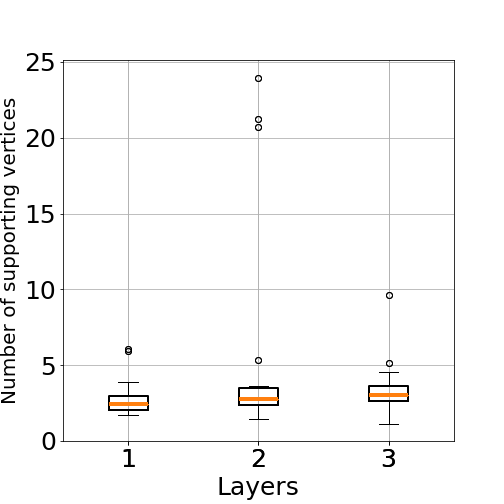}&
\includegraphics[width=.3\linewidth]{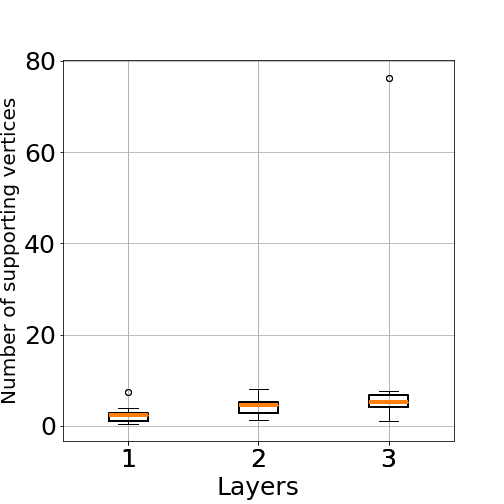}&
\includegraphics[width=.3\linewidth]{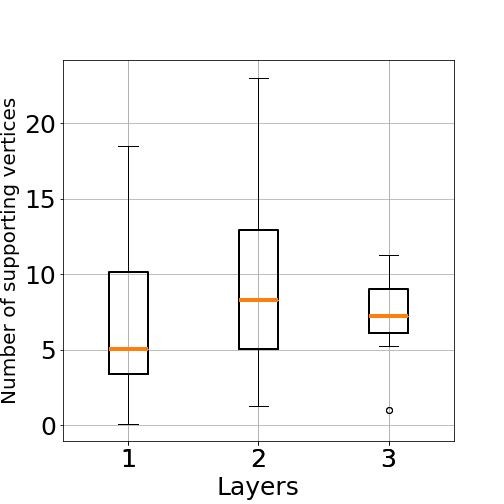}
\\
Mutagen & DD & Proteins\\
\end{tabular}
\caption{\modif{Boxplot of the number of supporting vertices per graph for layers 1, 2 and 3. \label{fig:bp_supp}}}
\end{center}
\end{figure}

\begin{figure}[htb]
\begin{center}
\begin{tabular}{ccc}
   \includegraphics[width=.273\linewidth]{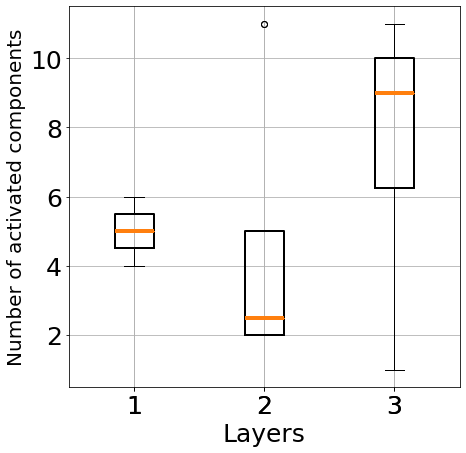}&
\includegraphics[width=.3\linewidth]{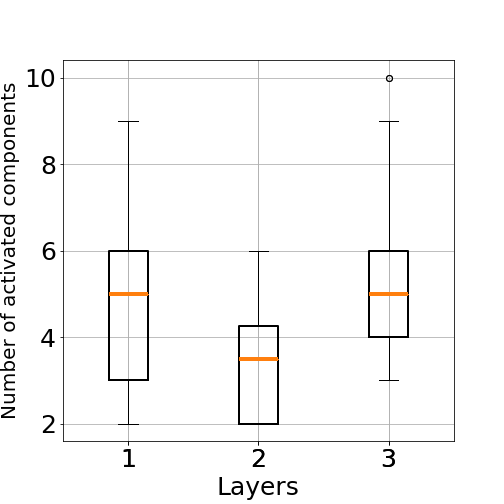}&
\includegraphics[width=.3\linewidth]{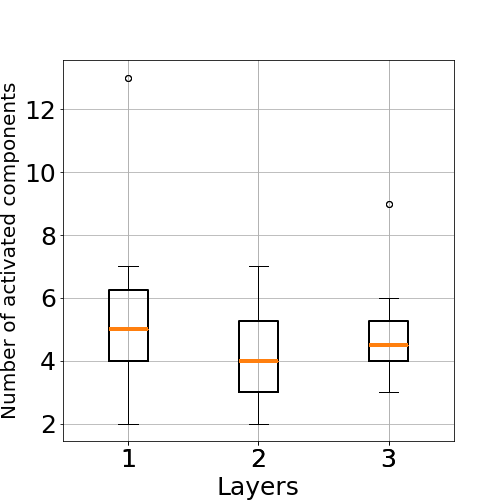}\\
BA2&  Aids & BBBP \\
\includegraphics[width=.3\linewidth]{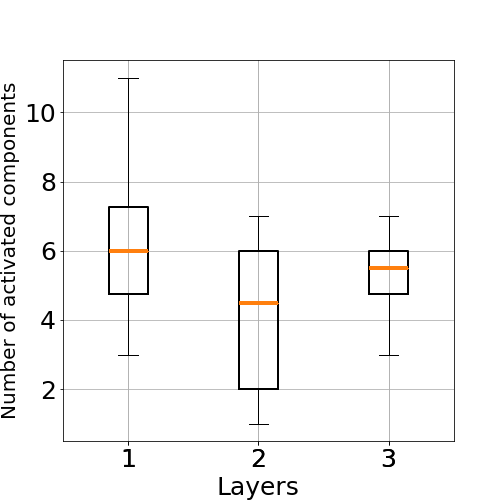}&
\includegraphics[width=.3\linewidth]{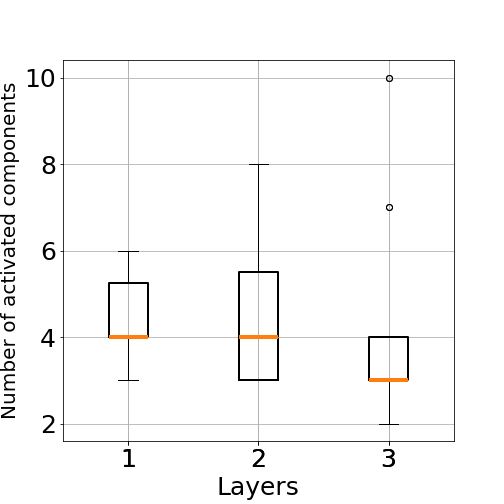}&
\includegraphics[width=.3\linewidth]{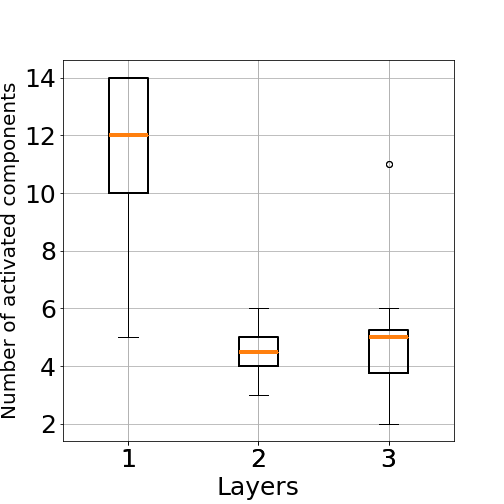}\\
Mutagen & DD & Proteins\\
\end{tabular}
\caption{\modif{Boxplot of the number of components per pattern for layers 1, 2 and 3.}
  \label{fig:bp_comp}}
\end{center}
\end{figure}

\begin{figure}[htb]
\begin{center}
\begin{tabular}{ccc}
   \includegraphics[width=.3\linewidth]{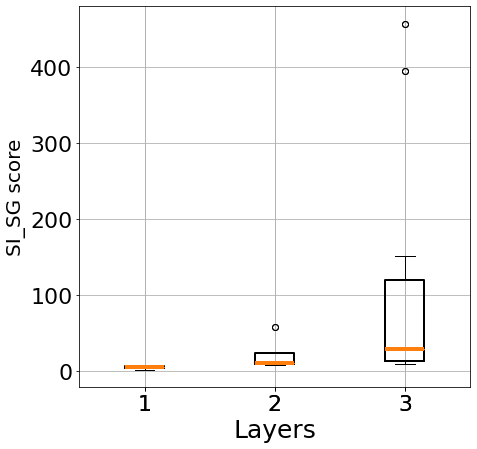}&
 \includegraphics[width=.3\linewidth]{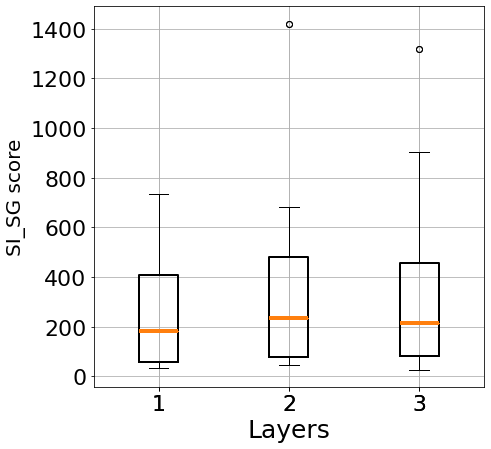}&
 \includegraphics[width=.3\linewidth]{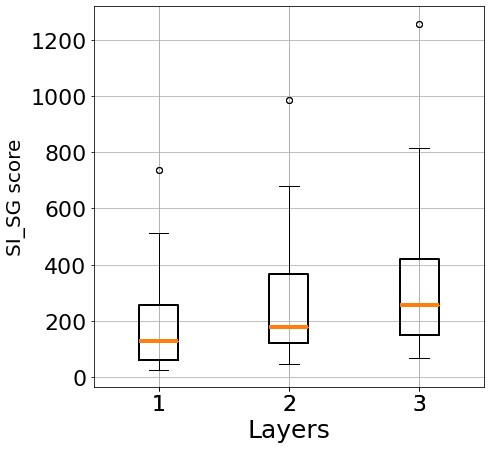}\\
  BA2&  Aids & BBBP \\
\includegraphics[width=.3\linewidth]{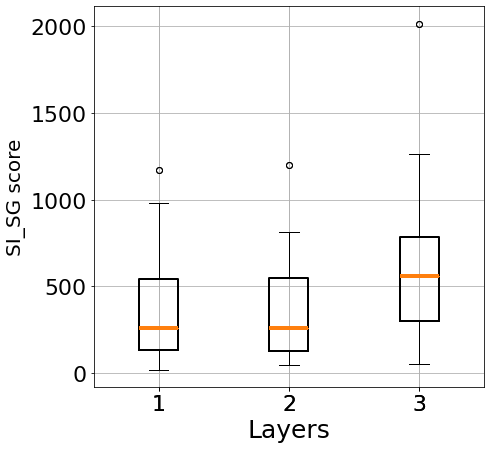}&
\includegraphics[width=.3\linewidth]{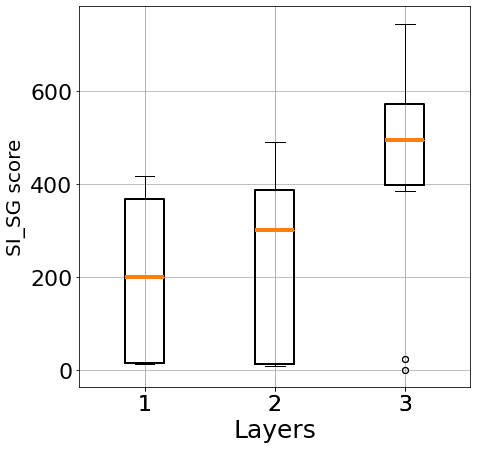}&
\includegraphics[width=.3\linewidth]{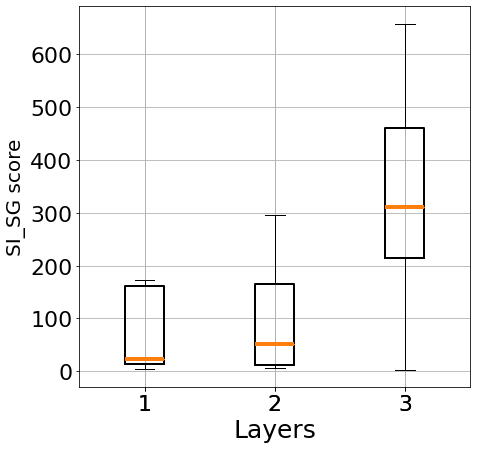}\\
Mutagen & DD & Proteins\\
\end{tabular}
\caption{Boxplot of the \modif{SI\_SG score of patterns for layers 1, 2 and 3.} \label{fig:bp_score}}
\end{center}
\end{figure}

\begin{figure}
\begin{center}
\begin{tabular}{cc}
     \includegraphics[width=.45\linewidth]{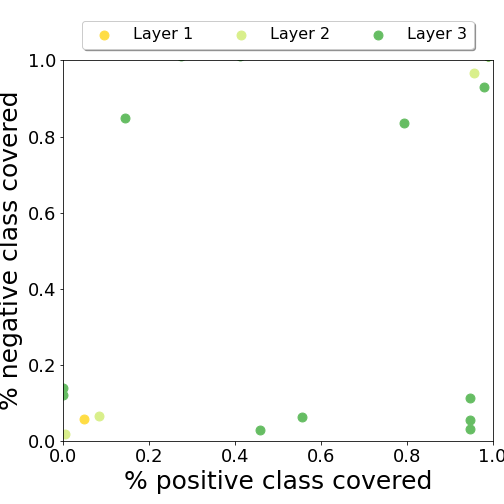}&
     \includegraphics[width=.45\linewidth]{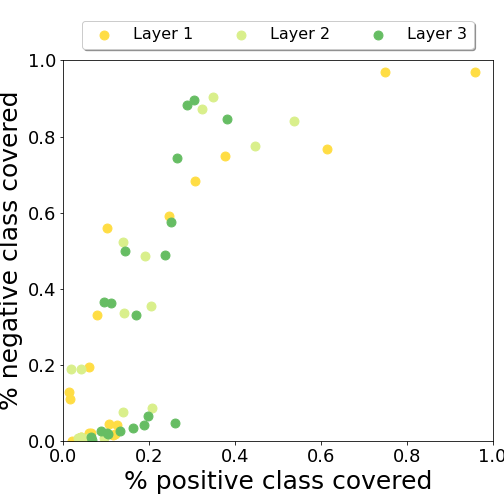}\\
      BA2&  Aids\\
   \includegraphics[width=.45\linewidth]{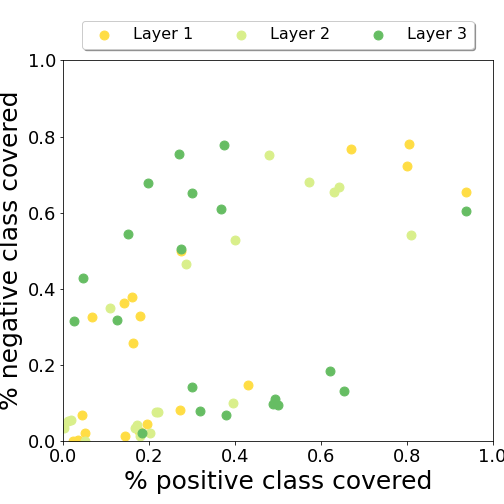}&
   \includegraphics[width=.45\linewidth]{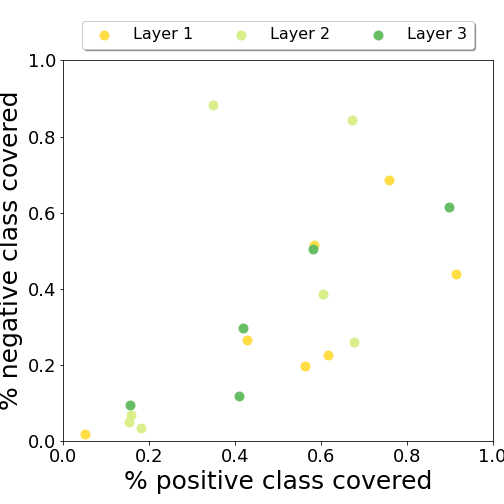}\\
    BBBP & Mutagen\\
   \includegraphics[width=.45\linewidth]{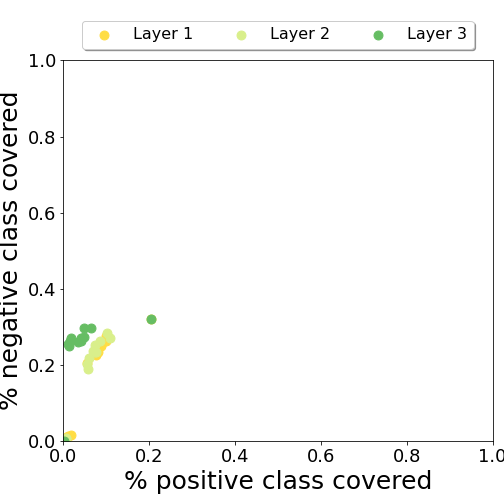}&
\includegraphics[width=.45\linewidth]{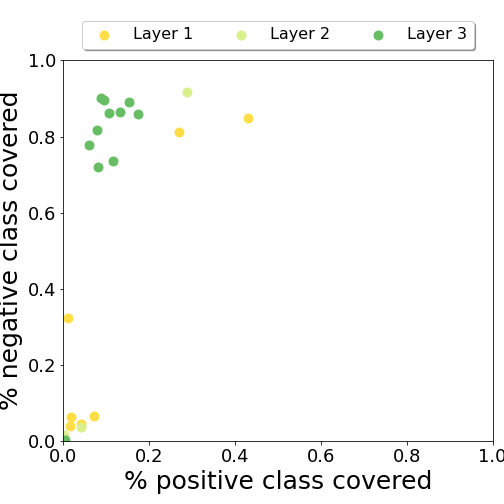}\\
DD & Proteins\\
\end{tabular}
\caption{\modif{Coverage of positive and negative classes, coloured according to the layers. A ``perfect'' discriminating pattern for the positive class (resp. negative class) would be projected to the
  lower right corner (resp. upper left corner).\label{fig:scp}}}
\end{center}
\end{figure}

The general characteristics of the activation \modif{rules} for each dataset are provided in Figs.~\ref{fig:bp_supp}--\ref{fig:scp}. One can observe -- in Fig. \ref{fig:bp_supp} --  that a rule is usually supported by more than one \modif{node} within a graph. Rules from the first layer of the GNN tend to involve a higher number of vertices than those in the following layers. It may be due to the fact that the first layer captures some hidden common features about the direct neighborhood of the vertices. The features captured by the GNN become more discriminant with layer indexes, as evidenced by the increasing SI\_SG score with layers in Fig. \ref{fig:bp_score}.
For some datasets (e.g., BA2, AIDS, DD, Proteins), some rules have high discriminative power for the positive class (bottom right corner in  Fig.~\ref{fig:scp}) or the negative class (top left corner). Their discriminative power is less effective for Mutagen and BBBP datasets. The most discriminant rules come from the last layer of the GNN.   Some rules are not discriminant (i.e., around the diagonal) but remains subjectively interesting. These rules uncover activated components that capture general properties of the studied graphs.
It is important to note that we study here the discriminative power of a rule according to its presence in graphs. These rules can be more discriminant if we take into account the number of occurrences of the rules  in the graphs. For instance, a rule that is not discriminant can becomes highly discriminant if we add a condition on its number of occurrences in graph, as we did when learning the decision trees in Table~\ref{tab:mim}.

%

\subsection{Comparison with competitors for explainability of GNN output}
We now assess the ability of activation \modif{rules} to provide good
explanations for the GNN decisions. According to the literature, the best competitors are
GNNExplainer \citep{YingBYZL19}, PGExplainer \citep{LuoCXYZC020} and PGM-Explainer \citep{VuT20}.
We  consider all of them as baseline methods. Furthermore, we also consider a gradient-based method \citep{PopeKRMH19}, denoted Grad,  even if it has been shown that such method is outperformed by the three others.   Therefore, we  compare \method{} against these 4 single-instance-explanation methods  in our experiments.

Evaluating the reliability of an explanation is not trivial due to the lack of ground truths. In our case, only BA2 is provided with ground truths by construction. When we have ground truths, we expect a good explanation to match it perfectly, but sometimes the model captures a different explanation that is just as discriminating.
Moreover, if fully present, ground truths contain only simple relationships (e.g., BA2) which are not sufficient for a full assessment.
Therefore, to be able to consider synthetic and real-world datasets, we consider a ground truth free metric.
\modif{We opt for {\em Fidelity} \citep{PopeKRMH19} which is defined as the difference of accuracy (or predicted probability) between the predictions on the original graph and the one obtained when masking part of the graph based on the explanations:}
$$Fid^{acc}= \frac{1}{N}\times \sum_{i=1}^N (1 - \delta_{(\hat{y}_i^{g_i \setminus m_i} = y_i)}),$$
where $y_i$ is the original prediction of graph $g_i$, $m_i$ is the mask and
$g_i \setminus m_i$ is the complementary mask, $\hat{y}_i^{g_i \setminus m_i}$ is the prediction for the complementary mask and $\delta_{(\hat{y}_i^{g_i \setminus m_i} = y_i)}$ equals 1 if both predictions are equal.

The fidelity can also be measured by studying the raw probability score given by the model for each class instead of the accuracy:
$$Fid^{prob}= \frac{1}{N}\times \sum_{i=1}^N (f(g_i)_{y_i}  - f(g_i \setminus m_i)_{y_i}),$$ with $f(g)_{y_i}$ is the prediction score for class $y_i$.

Similarly, we can study the prediction change by keeping important features (i.e., the mask)  and removing the others as Infidelity measures do:
$$\text{Infid}^{acc}=  \frac{1}{N}\times \sum_{i=1}^N (1 - \delta_{(\hat{y}_i^{m_i} = y_i)})$$  $$\text{Infid}^{prob}= \frac{1}{N}\times \sum_{i=1}^N (f(g_i)_{y_i}  - f(m_i)_{y_i}).$$
The higher the fidelity, the lower the infidelity, the better the explainer.

Obviously, masking all the input graph would have important impact to the model prediction. Therefore, the former measures should not be studied without considering the {\em Sparsity} metric that aims to measure the fraction of graph selected as mask by the explainer: $$ Sparsity= \frac{1}{N} \sum_{i=1}^{N}{\left(1 - \frac{|m_i|}{|g_i|}\right)},$$
where $|m_i|$ denotes the size of the mask $m_i$ and $|g_i|$ is the size of $g_i$ (the size includes the number of nodes, of edges and the attributes associated to them). Based on these measures, a better explainability method achieves higher fidelity, lower infidelity while keeping a sparsity  close to $1$.

We devise four policies to build a mask from an activation \modif{rules}:
\begin{itemize}
    \item[(1)] \textbf{node}: the simplest policy which takes only the nodes that are covered by the activation \modif{rule} and the edges adjacent to these nodes.
    \item[(2)] \textbf{ego}: the ego-graphs of radius $\ell$ centered on activated nodes, with $\ell$ the layer associated to the pattern.
\item[(3)] \textbf{decay}: a continuous mask with a weight associated to the edges that depends on the distance of its end-points to the activated nodes:
$$w_v =\sum_{a\in V_\mathcal{A}} \frac{1}{2^{1+d(v,a})} \text{ if }  d(v,a)\leq \ell,  0 \text{ otherwise}$$ with $V_\mathcal{A}$ the set of activated nodes, $d(v,a)$  the geodesic
distance between nodes $v$ and $a$
and $w_{(u,v)}=w_u + w_v$.
    \item[(4)] \textbf{top $k$}: a discrete mask containing only the $k$ edges from \textbf{decay} mask with the highest weights ($k=5$ or $k=10$ in our experiments).
\end{itemize}
For each policy, we select the mask (and the related pattern) that
maximises the fidelity. As GNNExplainer and PGExplainer provide
continuous masks,
we report, for fair comparisons,
the performance with both continuous and discrete masks built
with the $k$ best edges.
Note that the average time to provide an explanation ranges from 8ms to 84ms for \method.  This is faster than PGM-Explainer (about 5s), GNNExplainer (80ms to  240ms) and Grad (300ms). It remains slightly slower than PGExplainer (6ms to 20ms).
Table~\ref{tab:fid}(a) summarises the performance of the explainers
based on the Fidelity measures. Results show that \method{}
outperforms the baselines regardless of policy. On average, the gain
of our method against the best baseline is $231$\% for $Fid^{prob}$
and $207$\% for $Fid^{acc}$. These results must be analysed while
considering the sparsity (see Table~\ref{tab:fid}(c)).  In most of
the cases, \method{} provides sparser explanation than the
baselines. Furthermore, at equal sparsity (top $k$), \method{} obtains higher
fidelity values than both competitors. Notice that PGM-Explainer fails on BA2 because this dataset does not have labeled nodes and this method investigate only the nodes of the graphs.

We provide additional information on the Fidelity in Table~\ref{tab:fid_pol}. The Fidelity aims to measure the percentage of times that a model decision is changed when the input graphs is obfuscated by the mask m. In Table~\ref{tab:fid_pol}, we report a polarized version of the Fidelity for which we count the number of changes between the two possible decisions of the model. For instance, \modif{$F^{0 \to 1}$} measures the percentage of graphs initially classified  as `false' by the model that become classified as 'true' when obfuscating the graph with a mask. We can observe a dissymmetry between the class changes. As an example, \method{} has a perfect fidelity on BA2 and DD when considering only the positive examples, i.e., the mask provided by \method{} makes the model change its decision. When dealing with the negative examples, we obtain much lower score. Intuitively, some class changes  cannot be done by only removing some vertices or edges. Regarding BA2, it is impossible to obtain a house motif from a cycle without adding an edge to form a triangle.

The quality of the explanations are also assessed with the Infidelity metrics in
Table~\ref{tab:fid}(b). \method{} achieves excellent performance on BA2. On the other datasets, \method{} is outperformed by GNNExplainer. \method{} obtain similar scores or outperforms the other competitors  (i.e., PGExplainer, PGM-Explainer, Grad) at equal sparsity on most of the datasets.
Notice that, in these experiments, we made the choice to build mask based on a single activation \modif{rule} which is not enough to obtain fully discriminant mask for complex datasets. This is in agreement with what we observed in Fig. \ref{fig:scp}. We have no fully discriminant activation \modif{rule} for the positive and negative classes. Hence, it
would be necessary to combine activation \modif{rules} to build a more discriminant
mask and thus better optimise the Infidelity.


\begin{table}[!h]
  \begin{center}
\caption{Assessing the explanations with Fidelity, Infidelity and Sparsity metrics.} \label{tab:fid}
\rotatebox{90}{
\scalebox{0.82}{
\begin{tabular}{|l||c|c|c|c|c|c|c|c|c|c|c|c|}
\hline
{\bf (a) Fidelity}&\multicolumn{2}{c|}{DD}&\multicolumn{2}{c|}{Proteins}&\multicolumn{2}{c|}{BA2}&\multicolumn{2}{c|}{Aids}&\multicolumn{2}{c|}{BBBP}&\multicolumn{2}{c|}{Mutagen}\\

Model & $Fid^{prob}$  & $Fid^{acc}$ & $Fid^{prob}$& $Fid^{acc}$ & $Fid^{prob}$& $Fid^{acc}$ & $Fid^{prob}$ & $Fid^{acc}$  & $Fid^{prob}$  & $Fid^{acc}$  & $Fid^{prob}$ & $Fid^{acc}$  \\
\hline
\method(ego)     & 0.540 & 0.663 	 & 0.362 & 0.651 	 & 0.342 & 0.494 	& 0.165 & 0.097 	& 0.344 & 0.295  	& 0.492 & 0.647 \\
\method(node)     & 0.490 & 0.567 	 & 0.359 & 0.634 	 & 0.342 & 0.494 	& 0.175 & 0.076 	& 0.362 & 0.336  	& 0.582 & 0.833 \\
\method(decay)   & 0.447 & 0.485 	 & 0.344 & 0.576 	 & 0.342 & 0.494 	& 0.145 & 0.055 	& 0.316 & 0.276  	& 0.554 & 0.781 \\
\method(top 5)   & 0.276 & 0.421 	 & 0.069 & 0.086 	 & 0.353 & 0.917 	& 0.160 & 0.058 	& 0.271 & 0.260  	& 0.450 & 0.629 \\
\method(top 10)  & 0.296 & 0.445 	 & 0.092 & 0.127 	 & 0.220 & 0.496 	& 0.160 & 0.057 	& 0.304 & 0.270  	& 0.458 & 0.600 \\
\hline
Grad            & 0.083 & 0.089 	 & 0.060 & 0.084 	& 0.195 & 0.494 	& 0.078 & 0.018  	& 0.171 & 0.132  	& 0.223 & 0.254\\
GnnExplainer    & 0.077 & 0.086 	 & 0.021 & 0.037 	& 0.093 & 0.198 	& 0.036 & 0.009  	& 0.100 & 0.101  	& 0.177 & 0.227\\
PGExplainer  	& 0.070 & 0.082 	 & 0.019 & 0.034 	& 0.004 & 0.000 	& 0.032 & 0.010  	& 0.098 & 0.099  	& 0.157 & 0.179\\
Grad(top5)      & 0.080 & 0.085 	 & 0.042 & 0.081 	& 0.087 & 0.175 	& 0.059 & 0.013  	& 0.126 & 0.107 	& 0.222 & 0.263\\
GnnExplainer(top5)    & 0.020 & 0.027 	 & 0.026 & 0.053 	& 0.183 & 0.461 	& 0.060 & 0.018  	& 0.086 & 0.079 	& 0.226 & 0.305\\
PGExplainer(top5)  	& 0.021 & 0.027 	 & 0.038 & 0.058 	& 0.182 & 0.516 	& 0.066 & 0.019  	& 0.148 & 0.125 	& 0.199 & 0.236\\
Grad(top 10)        & 0.083 & 0.089 	& 0.060 & 0.084  	& 0.195 & 0.494 	& 0.078 & 0.018 	 & 0.171 & 0.132 	& 0.223 & 0.254 \\
GnnExplainer(top 10)  	& 0.034 & 0.042 	& 0.043 & 0.088  	& 0.200 & 0.491 	& 0.074 & 0.018 	 & 0.125 & 0.104 	& 0.293 & 0.400 \\
PGExplainer(top 10)  	& 0.032 & 0.036 	& 0.046 & 0.072  	& 0.206 & 0.517 	& 0.083 & 0.030 	 & 0.165 & 0.117 	& 0.206 & 0.258 \\
PGM-Explainer  & 0.233 & 0.339 	& 0.096 & 0.207 & 0.000 & 0.000 	& 0.089 & 0.028 	& 0.212 & 0.198  	& 0.260 & 0.338 \\
\hline\hline
{\bf (b) Infidelity} & $\text{Infid}^{prob}$  & $\text{Infid}^{acc}$ &$\text{Infid}^{prob}$  & $\text{Infid}^{acc}$ &$\text{Infid}^{prob}$  & $\text{Infid}^{acc}$ & $\text{Infid}^{prob}$   &$\text{Infid}^{acc}$  & $\text{Infid}^{prob}$  & $\text{Infid}^{acc}$  & $\text{Infid}^{prob}$  & $\text{Infid}^{acc}$ \\
\hline

\method(ego)     & 0.133 & 0.062 	& 0.163 & 0.188 	 & 0.000 & 0.000 &  	0.766 & 0.806 	& 0.369 & 0.452 	& 0.273 & 0.349\\
\method(node)     & 0.133 & 0.048 	& 0.160 & 0.196 	 & 0.000 & 0.000 &  	0.767 & 0.806 	& 0.374 & 0.464 	& 0.237 & 0.288\\
\method(decay)   & 0.140 & 0.097 	& 0.162 & 0.202 	 & 0.000 & 0.000 &  	0.767 & 0.806 	& 0.362 & 0.454 	& 0.233 & 0.272\\
\method(top 5)   & 0.341 & 0.340 	& 0.287 & 0.355 	 & 0.323 & 0.494 &  	0.770 & 0.806 	& 0.441 & 0.574 	& 0.341 & 0.460\\
\method(top 10)  & 0.341 & 0.340 	& 0.297 & 0.355 	 & 0.310 & 0.494 &  	0.768 & 0.806 	& 0.405 & 0.524 	& 0.329 & 0.435\\

\hline

Grad  & 0.344 & 0.340  	& 0.326 & 0.355 	& 0.334 & 0.494		& 0.769 & 0.806		& 0.447 & 0.623  	& 0.357 & 0.489 \\
GnnExplainer  	& 0.075 & 0.084  	& 0.021 & 0.036 	& 0.223 & 0.494		& 0.036 & 0.012		& 0.099 & 0.098  	& 0.140 & 0.141 \\
PGExplainer  	& 0.082 & 0.086  	& 0.024 & 0.039 	& 0.353 & 0.494		& 0.038 & 0.012		& 0.098 & 0.096  	& 0.157 & 0.185 \\
Grad(top 5)  & 0.343 & 0.340  	& 0.312 & 0.355 	& 0.327 & 0.494		& 0.770 & 0.806		& 0.471 & 0.651  	& 0.356 & 0.485 \\
GnnExplainer(top 5)  	& 0.348 & 0.498  	& 0.228 & 0.599 	& 0.321 & 0.494		& 0.101 & 0.057		& 0.216 & 0.179  	& 0.297 & 0.354 \\
PGExplainer(top 5)  	& 0.343 & 0.340  	& 0.296 & 0.355 	& 0.332 & 0.494		& 0.769 & 0.806		& 0.510 & 0.695  	& 0.353 & 0.490 \\
Grad(top 10)  & 0.344 & 0.340  	& 0.326 & 0.355 	& 0.334 & 0.494		& 0.769 & 0.806		& 0.447 & 0.623  	& 0.357 & 0.489 \\
GnnExplainer(top 10)  	& 0.343 & 0.474  	& 0.197 & 0.491 	& 0.308 & 0.494		& 0.105 & 0.054		& 0.206 & 0.180  	& 0.282 & 0.343 \\
PGM-Explainer  & 0.345 & 0.340 	& 0.341 & 0.355 & 0.342 & 0.494 	& 0.765 & 0.806 	& 0.392 & 0.514  	& 0.354 & 0.498	\\

\hline
\hline

{\bf (c) Sparsity}&\multicolumn{2}{c|}{DD}&\multicolumn{2}{c|}{Proteins}&\multicolumn{2}{c|}{BA2}&\multicolumn{2}{c|}{Aids}&\multicolumn{2}{c|}{BBBP}&\multicolumn{2}{c|}{Mutagen}\\\hline

\method(ego)  		& \multicolumn{2}{c|}{0.544} & \multicolumn{2}{c|}{0.410} & \multicolumn{2}{c|}{0.011} & \multicolumn{2}{c|}{0.822} & \multicolumn{2}{c|}{0.805} & \multicolumn{2}{c|}{0.717}\\
\method(node)  		& \multicolumn{2}{c|}{0.769} & \multicolumn{2}{c|}{0.429} & \multicolumn{2}{c|}{0.002} & \multicolumn{2}{c|}{0.897} & \multicolumn{2}{c|}{0.870} & \multicolumn{2}{c|}{0.731}\\
\method(decay)  	& \multicolumn{2}{c|}{0.717} & \multicolumn{2}{c|}{0.394} & \multicolumn{2}{c|}{0.010} & \multicolumn{2}{c|}{0.870} & \multicolumn{2}{c|}{0.860} & \multicolumn{2}{c|}{0.697}\\
\method(top 5)  	& \multicolumn{2}{c|}{0.997} & \multicolumn{2}{c|}{0.993} & \multicolumn{2}{c|}{0.902} & \multicolumn{2}{c|}{0.955} & \multicolumn{2}{c|}{0.969} & \multicolumn{2}{c|}{0.989}\\
\method(top 10)  	& \multicolumn{2}{c|}{0.994} & \multicolumn{2}{c|}{0.986} & \multicolumn{2}{c|}{0.804} & \multicolumn{2}{c|}{0.915} & \multicolumn{2}{c|}{0.939} & \multicolumn{2}{c|}{0.978}\\
\hline

Grad  & \multicolumn{2}{c|}{0.994} & \multicolumn{2}{c|}{0.986} & \multicolumn{2}{c|}{0.804} & \multicolumn{2}{c|}{0.910} & \multicolumn{2}{c|}{0.938} & \multicolumn{2}{c|}{0.978}\\
GnnExplainer  	& \multicolumn{2}{c|}{0.502} & \multicolumn{2}{c|}{0.501} & \multicolumn{2}{c|}{0.619} & \multicolumn{2}{c|}{0.501} & \multicolumn{2}{c|}{0.501} & \multicolumn{2}{c|}{0.505}\\
PGExplainer  	& \multicolumn{2}{c|}{0.529} & \multicolumn{2}{c|}{0.545} & \multicolumn{2}{c|}{0.955} & \multicolumn{2}{c|}{0.547} & \multicolumn{2}{c|}{0.534} & \multicolumn{2}{c|}{0.515}\\
PGM-Explainer  & \multicolumn{2}{c|}{0.973}& \multicolumn{2}{c|}{0.955} & \multicolumn{2}{c|}{nan} & \multicolumn{2}{c|}{0.855} & \multicolumn{2}{c|}{0.884} & \multicolumn{2}{c|}{0.956}\\

\hline
\end{tabular}
}
}
\end{center}
\end{table}

\begin{table}[!ht]
\begin{center}
  \caption{Polarized fidelity.}
  \label{tab:fid_pol}
\rotatebox{90}{
\scalebox{0.9}{
\begin{tabular}{|l||c|c|c|c|c|c|c|c|c|c|c|c|}
\hline
{\bf (a) Fidelity}&\multicolumn{2}{c|}{DD}&\multicolumn{2}{c|}{Proteins}&\multicolumn{2}{c|}{BA2}&\multicolumn{2}{c|}{Aids}&\multicolumn{2}{c|}{BBBP}&\multicolumn{2}{c|}{Mutagen}\\
Model & \modif{$F^{0\to 1}$}  & $F^{1\to 0}$ & $F^{0\to 1}$  & $F^{1\to 0}$ &$F^{0\to 1}$  & $F^{1\to 1}$ &$F^{0\to 1}$  & $F^{1\to 0}$ &$F^{0\to 1}$  & $F^{1\to 0}$ &$F^{0\to 1}$  & $F^{1\to 0}$ \\
\hline
\method(ego)     & 0.489 & 1.000 &	 0.572 & 0.795 &	0.000 & 1.000 &	    0.353 & 0.035 &   0.965 & 0.137 & 0.543 & 0.809\\
\method(node)     & 0.344 & 1.000 &	 0.546 & 0.795 &	0.000 & 1.000 &	    0.198 & 0.047 &   0.981 & 0.184 & 0.795 & 0.891\\
\method(decay)   & 0.219 & 1.000 &	 0.455 & 0.795 &	0.000 & 1.000 &	    0.180 & 0.025 &   0.933 & 0.121 & 0.744 & 0.838\\
\method(top 5)   & 0.135 & 0.977 &	 0.029 & 0.190 &	0.836 & 1.000 &	    0.080 & 0.053 &   0.808 & 0.130 & 0.500 & 0.830\\
\method(top 10)  & 0.167 & 0.985 &	 0.045 & 0.276 &	0.004 & 1.000 &	    0.098 & 0.047 &   0.869 & 0.129 & 0.398 & 0.913\\

\hline
Grad                    & 0.091 & 0.086     & 0.022 & 0.195     & 0.000 & 1.000     & 0.046 & 0.011     & 0.569 & 0.029      & 0.053 & 0.567\\
GnnExplainer            & 0.031 & 0.191     & 0.018 & 0.071     & 0.000 & 0.401     & 0.026 & 0.005     & 0.495 & 0.008      & 0.097 & 0.429\\
PGExplainer             & 0.029 & 0.186     & 0.017 & 0.066     & 0.000 & 0.000     & 0.023 & 0.007     & 0.511 & 0.002      & 0.072 & 0.345\\
Grad(top 5)             & 0.087 & 0.081     & 0.040 & 0.154     & 0.002 & 0.352     & 0.039 & 0.007     & 0.454 & 0.026      & 0.240 & 0.299\\
GnnExplainer(top 5)     & 0.013 & 0.055     & 0.026 & 0.101     & 0.049 & 0.883     & 0.034 & 0.014     & 0.265 & 0.035      & 0.265 & 0.368\\
PGExplainer(top 5)      & 0.010 & 0.060     & 0.053 & 0.068     & 0.423 & 0.611     & 0.098 & 0.001     & 0.581 & 0.017      & 0.251 & 0.214\\
Grad(top 10)            & 0.091 & 0.086     & 0.022 & 0.195     & 0.000 & 1.000     & 0.046 & 0.011     & 0.569 & 0.029      & 0.053 & 0.567\\
GnnExplainer(top 10)    & 0.023 & 0.078     & 0.033 & 0.187     & 0.000 & 0.994     & 0.067 & 0.006     & 0.419 & 0.030      & 0.317 & 0.527\\
PGExplainer(top 10)     & 0.012 & 0.083     & 0.070 & 0.076     & 0.077 & 0.968     & 0.155 & 0.000     & 0.556 & 0.014      & 0.212 & 0.331\\
PGM-Explainer            & 0.198 & 0.612 	& 0.109 & 0.385     & 0.000 & 0.000 	& 0.111 & 0.008 	& 0.645 & 0.093   	& 0.130 & 0.662\\

\hline

\end{tabular}
}
}
\end{center}
\end{table}

\subsection{Model insights via the (re)description of activation patterns} 
We argue that activation \modif{rules} also help provide insight into the
model, especially what the GNN model captures.  As discussed in Section \ref{sec:charact}, this requires characterizing the nodes (and their neighborhood) that support a given activation \modif{rule}. In this experimental study, we investigate the obtained numerical subgroups for BA2 and the subgraph characterizing the activation \modif{rules} retrieved for Mutagen, BBBP and Aids datasets.

\subsubsection{Numerical subgroups} 
\begin{figure}[htb]
\begin{center}
\begin{tabular}{cc}
	\includegraphics[width=.35\linewidth]{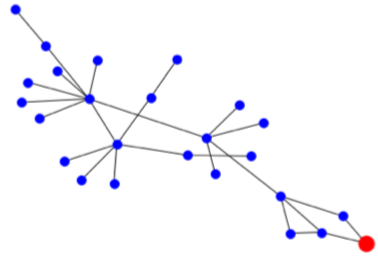}&
	\includegraphics[width=.35\linewidth]{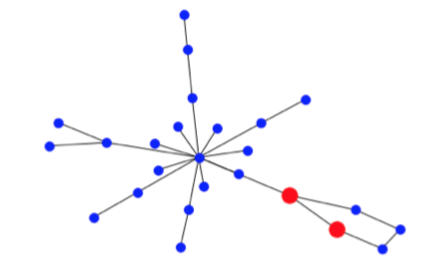}\\
	\texttt{\small clustering=0.0 AND degree=2 } & \texttt{clustering2=0.0 AND degree2: [7:8[  } \\
	\texttt{AND triangle2=1 AND triangle=0} & \texttt{WRAcc=0.12 }\\
	\texttt{WRAcc=0.24} & \texttt{}\\
\end{tabular}
\caption{Nodes (in red) in the support of two activation \modif{rules} that are discriminant for $p^1$ support, related to the positive target (left), and for $p^0$ support, related to the negative target (right).} \label{fig:topo-ba2}
\end{center}
\end{figure}

Each node can be easily described
with some topological properties (e.g., its degree, the number of
triangles it is involved in). Similarly, we can describe its
neighborhood by aggregating the values of the neighbors.  Thanks to
such properties, we make a propositionalization of the nodes of the graphs. Considering the two most discriminant activation \modif{rules}\footnote{\modif{$p^1=\{a_3,
a_6, a_7, a_9, a_{10},a_{15}\}$ (where $a_i$ are the activated components of the rule and $p$ is the set representation of the bitset $A^\ell$ of Definition~\ref{def2}), $|\supp(p^1,{\cal
  G}^1)|=474$, $|\supp(p^1,{\cal G}^0)|=16$ and $p^0=\{a_{0},a_{1},a_{2},a_{4},a_{5},a_{8},a_{11},a_{17},a_{18},
a_{19}\}$, $|\supp(p^0,{\cal G}^1)|=137$, $|\supp(p^0,{\cal
  G}^0)|=506$.}}, we use
%
the subgroup discovery algorithm from pysubgroup library \cite{DBLP:conf/pkdd/Lemmerich018} to find the
discriminating conditions of the nodes supporting these
two patterns.
Fig. \ref{fig:topo-ba2} reports a visualisation of two graphs with activated nodes in red. The best description
based on WRAcc measure of pattern $p^1$ (Fig. \ref{fig:topo-ba2} left) and $p^0$ (Fig. \ref{fig:topo-ba2} right) are given below.
%
For the House motif (positive class of BA2), the
nodes that support activation \modif{rules} are almost perfectly described
(the WRacc equals to 0.24 while maximum value is 0.25) with the
following conditions: {\em Nodes connected to two neighbors
  (degree=2) that are not connected between them (clustering
  coefficient=0), not involved in a triangle and one of its neighbors
  is involved in a triangle (triangle2=1).}  In other words, the
activation \modif{rule} captures one node of the floor of the ``house
motif".  We have similar conditions to identify some nodes of the
$5$-node cycle (negative class of BA2): {\em nodes without triangle in their direct neighborhood (clustering2=0) and whose sum
  of neighbors' degree (including itself) equal $7$
  (degree2 $\in$[7:8[)}.

\begin{table}[htb]
\begin{center}
\caption{Characterization of activation \modif{rules} with numerical subgroups on  BA2. We only report the subgroup whose WRAcc value is greater than 0.1.}\label{tab:sgBA2}
\scalebox{0.99}{
\begin{tabular}{|c|c|p{8cm}|c|}
\hline
{\bf Layer}	& {\bf Class}	& {\bf Description}	& {\bf WRAcc}	\\ \hline \hline
2 & 0 & \texttt{degree=3} & 0.2475\\\hline
2 & 1 & \texttt{clustering2=0 AND  degree=2 AND triangle2\_avg=0 } & 0.207 \\
2 & 1 & \texttt{betweenness: [0.0:0.00[ AND clustering2=0.0} & 0.127\\\hline \hline
3 & 0 & \texttt{clustering2=0.0 AND degree2: [7:8[ AND degree2\_avg: [3.50:3.57[}  & 0.114 \\
3 & 0 & \texttt{clustering2=0.0 AND  degree=2 AND triangle2=0 }  &0.101 \\
3 & 0 & \texttt{betweenness2: [0.37:0.38[ AND betweenness2\_avg: [0.19:0.20[ AND clustering2=0.0}  & 0.202\\
3 & 0 & \texttt{betweenness2: [0.37:0.39[ AND betweenness2\_avg: [0.19:0.21[ AND betweenness=0.07608695652173914}  &0.209 \\
3 & 0 & \texttt{betweenness: [0.29:0.30[ AND clustering2=0.0 AND degree==3 }  & 0.147\\
3 & 0 & \texttt{betweenness: [0.0:0.00[ AND clustering2=0.0  AND degree2\_avg: [4.0:4.17[}  &0.162 \\ \hline
3 & 1 & \texttt{clustering=0.0 AND degree=2 AND triangle2\_avg=0.5}  &0.227 \\
3 & 1 & \texttt{degree2: [7:8[ AND degree2\_avg: [3.50:3.60[ AND degree=2 AND triangle=0}  &0.224 \\
3 & 1 & \texttt{degree=2 AND triangle2=1}  & 0.238\\
3 & 1 & \texttt{clustering==0.0 AND degree==2 AND triangle2==1 AND triangle==0}  &0.240 \\
3 & 1 & \texttt{degree=2}  & 0.125\\
3 & 1 & \texttt{clustering=0.0 AND degree=2 AND triangle2=1 AND triangle2\_avg=0.5}  & 0.232\\ \hline
\end{tabular}
}
\end{center}
\end{table}

We report the description in terms of numerical subgroups of the activation \modif{rules} in Table~\ref{tab:sgBA2}. It is important to note that even if some activation \modif{rules} were found as subjectively interesting according to a specific output of the model, they may capture some general properties of the  BA2 graph that are not so specific of one of the classes. For instance, the second subgroup is related to the positive class (i.e., house motif) but what it captured is not specific to house motif (degree=2, absence of triangle).





\subsubsection{\modif{Graph subgroups}}

\begin{figure}[htb]
\begin{center}
  \begin{tabular}{m{1cm}m{5cm}m{5cm}}
Aids    & \includegraphics[width=\linewidth]{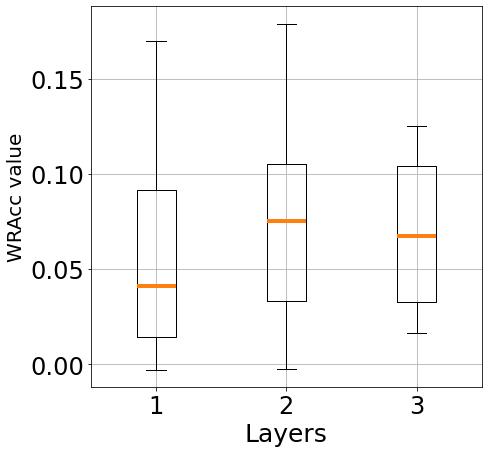}&\includegraphics[width=\linewidth]{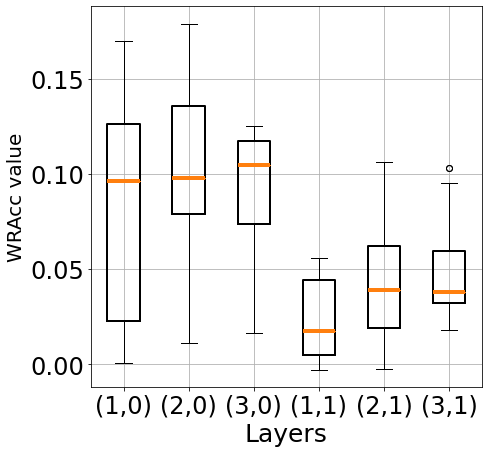}\\
BBBP & \includegraphics[width=\linewidth]{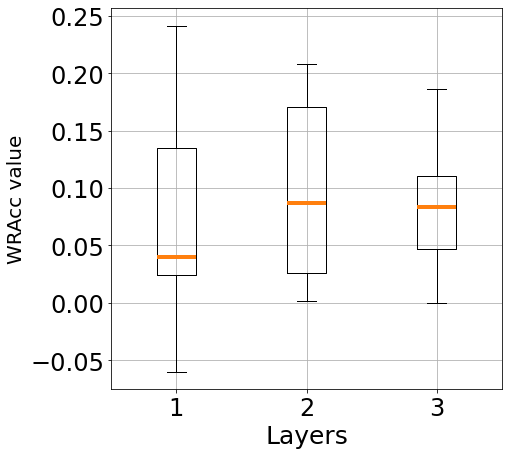}&\includegraphics[width=\linewidth]{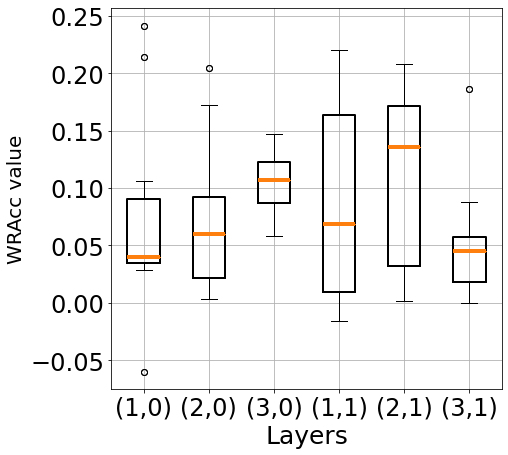}\\
Mutagen & \includegraphics[width=\linewidth]{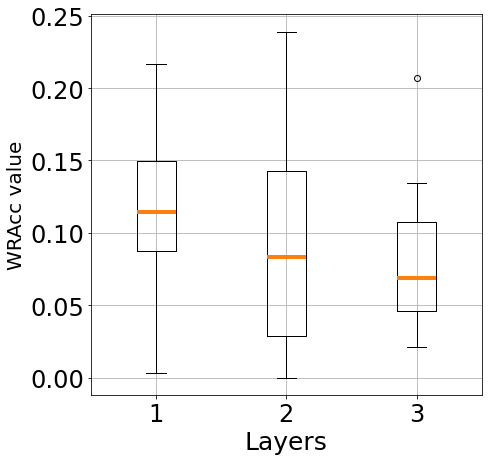}&\includegraphics[width=\linewidth]{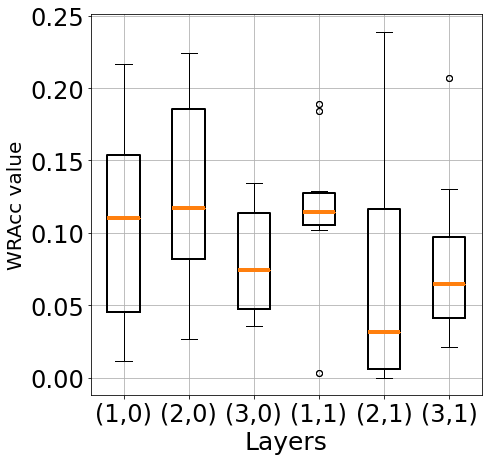}
  \end{tabular}

\end{center}
\caption{Boxplot of the WRAcc values of graph subgroups related to activation \modif{rules} by layer (left column) or by both layer and model decision (right column) for Aids (first row), BBBP (second row) and Mutagen (third row).}
\label{fig:wracc_graphs}
\end{figure}

Similarly, we can characterize activation \modif{rules} with \modif{graph subgroups}.
We investigate the interest of such pattern language for three datasets: Aids, BBBP and Mutagen.
\modif{For each activation rule, we compute the graph that has the maximum WRAcc value, using $\min\_sup=10$ (see Equation~\ref{eq1}). In other words, this graph has an important number of isomorphisms with ego-graphs that support the rule and that correspond to the class of the target of the rule.}
\modif{In Fig. \ref{fig:wracc_graphs}, we report the WRAcc values of the discovered graphs that aim to characterize the activation \modif{rules}. We can observe that the WRAcc values are rather high (WRAcc ranges from -1 to $0.25$) which demonstrates that these graphs well describe the parts of the GNN identified by the activation \modif{rules}.}


\begin{figure}[htb]
\begin{center}
\includegraphics[width=.9\linewidth]{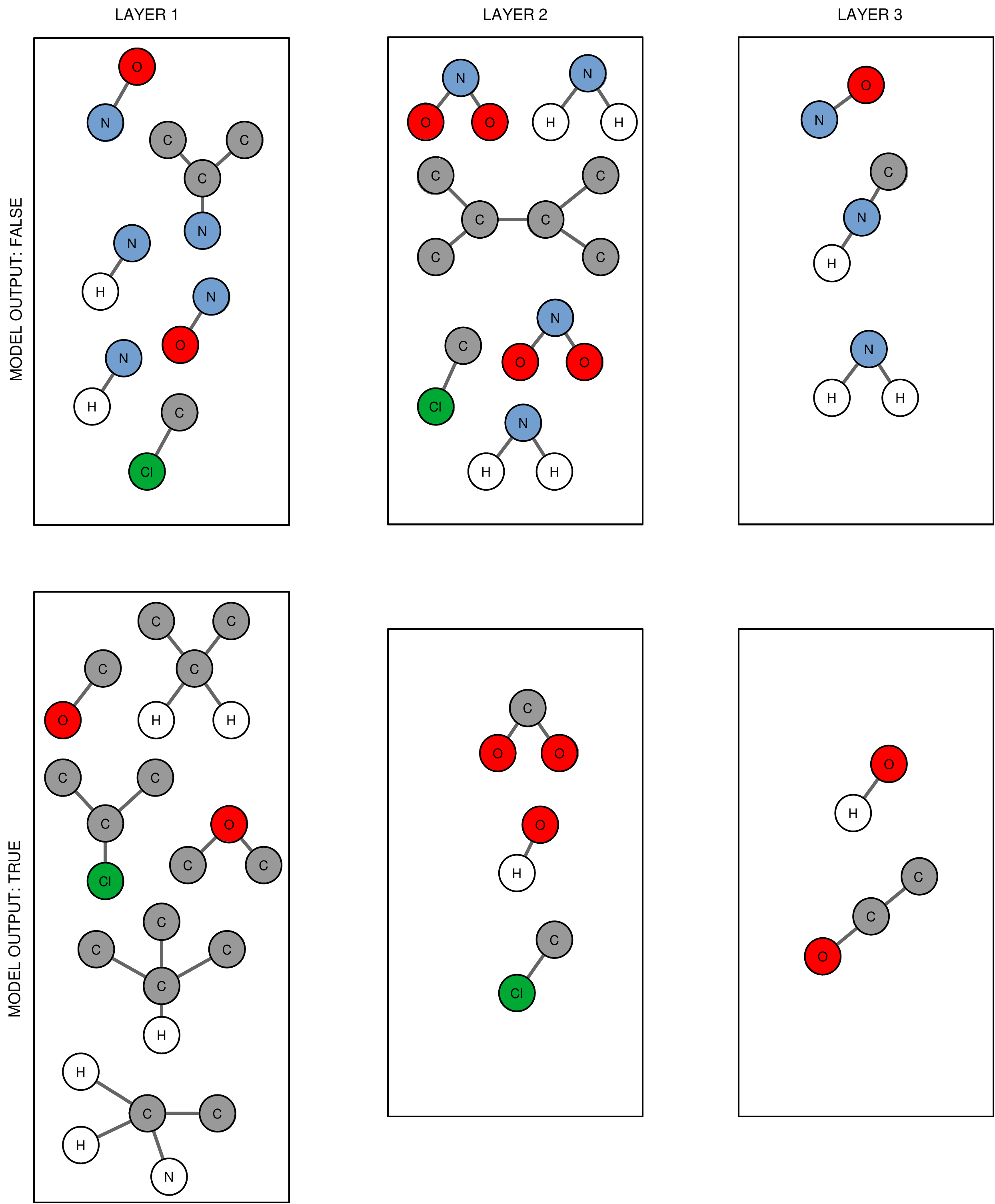}
\end{center}
\caption{Characterization of activation \modif{rules} for Mutagen with discriminant subgraphs. We retain only the subgraphs with a WRAcc value greater than $0.1$. Mutagenic chemicals are classified as False.}\label{fig:mutagen_summary}
\end{figure}

The subgraphs obtained for Mutagen dataset are summarised in Fig.~\ref{fig:mutagen_summary}. For each layer and decision, we display the subgraphs whose WRAcc is greater than 0.1 layer by layer.
The negative class is related to mutagenic molecules.  Several things can be observed from this figure. First, some subgraphs are known as toxicophores or fragment of toxicophores in  the literature \citep{kazius2005derivation}. For instance, the subgraph with two hydrogen and one azote atoms is a part of an aromatic amine. Similarly, the subgraph with one azote and two oxygen atoms is an aromatic nitro.  The subgraph involving 6 carbon atoms is a fragment of a bay region or a k-region.
Second, some subgraphs appear several times. It means that several activation rules are described with the same subgraphs. This can be explained in several ways. Neural networks are known to have a lot of redundant information, as evidenced by the numerous papers in the domain that  aim to compress or simplify deep neural networks \citep{chen2018constraint,pan2016dropneuron,pasandi2020modeling,xu2018deep}. Accordingly, this is not surprising to have several parts of the GNN that are similar and described by the same subgraphs. Notice that this problem could be an interesting perspective for our work.
Another explanation is that the subgraphs well describe the hidden features captured by the GNN but from different perspective, i.e., the center is different. For instance, for a simple chemical bond C-N, one may have the same graph with one centered in C and the other in N.
A last explanation could be that the subgraph language is not enough powerful to capture the subtle differences between the activation \modif{rules}. Once again, the definition of more sophisticated and appropriate languages to describe the hidden features captured by the GNN is a promising perspective of research.
%

These latter experiments show that \method{} represents a valuable alternative to GNN explainability methods. In addition to providing single instance explanations, \method{} can provide insights about what the GNN perceives. Especially, it allows to build a summary of the hidden features captured  by the model (e.g., Fig.~\ref{fig:mutagen_summary}). In relation to this, our method is quite analogous to model explanation methods such as XGNN \citep{XGNN}. This deserves a discussion and a comparison with XGNN.

\subsubsection{Comparison to XGNN}
XGNN \citep{XGNN} is a method rooted in reinforcement learning that generates graphs that maximise the model decision for a given class. For Mutagen, we generate 20 graphs for each class with a maximum size equal to 6. Considering the 40 generated graphs, we observe that only one of them is a subgraph of at least one graph of the dataset. The other graphs have on average 60\% of partial inclusion: the maximum common subgraph with molecules from Mutagen uncovers 60\% of a generated graph. Therefore, we can conclude that XGNN generates graphs that are not enough realistic. The only graph that appears within the dataset involves a carbon atom bonded to 2 others carbon atoms and one hydrogen atom. With \method{},
we obtained two subgraphs characterizing some activation \modif{rules} that are super-graphs of this one (see Fig.~\ref{fig:mutagen_summary}). Notice that, we also found this subgraph for some activation \modif{rules}. We did not report it in Fig. \ref{fig:mutagen_summary} because its WRAcc value is lower than 0.1. Nevertheless,  this graph appears in $21100$ ego-graphs in the dataset. It describes a fragment of molecule that is very common. One can wonder if such a fragment can be mutagenic or if XGNN has just captured it a biased of the GNN.
Furthermore, XGNN has generated graphs that are not planar, which is not common in Chemistry.
Based on these evidences, we argue that XGNN does not return realistic graphs while our approach -- by construction --  provides subgraphs from the dataset.

We search for each pattern produced by \method{} the closest pattern in XGNN according to the Graph Edit
Distance (GED) and vice versa.  We note that the previously described prototype graph (i.e., 3 carbons and 1 hydrogen) is found in most of the cases as being  the closest to the patterns produced by \method.
In average, the distance between each XGNN prototype and the closest pattern of \method{} is 4.6 while the mean distance between \method{} subgraphs and the closest from XGNN is 3.7. This is rather important since the graphs provided by XGNN  have at most $6$ nodes.

We believe that a model decision for a class cannot be summarized into a single prototype. Several different  phenomena can lead to the same class. Furthermore, as we observed, this can lead to unrealistic prototype even if domain knowledge is integrated within the graph generation. \method{} allows to have deeper insights from the GNN by considering each hidden feature separately.

%



\section{Discussion and Conclusion} 
We have introduced a novel method for the explainability of GNNs. \method{} is based on the discovery of relevant activation rules in each hidden layer of the GNN. Prior beliefs are used to assess how contrastive a rule is. We have proposed an algorithm that efficiently and iteratively builds a set of activation rules, limiting the redundancy between them.  Extensive empirical results on several real-world datasets confirm that the activation rules capture interesting insights about how the internal representations are built by the GNN. Based on these rules, \method{} outperforms the SOTA methods for GNN explainability \modif{when considering Fidelity metric}. Furthermore, the consideration of pattern languages involving interpretable features (e.g., numerical subgroups on node topological properties, graph subgroups) is promising since it makes possible to summarise the hidden features built by the GNN through its different layers. 

\modif{
We believe that such method can support knowledge discovery from powerful GNNs and provide insights  on object of study for scientists or more generally  for any user.
However, a number of potential limitations need to be considered for future research to make this knowledge discovery from GNNs effective in practice. }

\modif{First, assessing explanations without ground truth is not trivial. Our experimental evaluation relies on Fidelity, Infidelity and Sparsity metrics. Fidelity assumes that the GNN decision would change if key part of the graphs are removed. However, it is not always the case in practice. For instance, it is difficult to obtain a toxic molecule from a non-toxic one by only removing some atoms. That would be interesting to investigate other evaluation measures that take into account the negation (i.e., absence of important features) and evaluation measures based on the addition of subgraphs.}

\modif{In this paper, we  have devised  an exhaustive algorithm for discovering the activation rules. Even if pruning  based on upper bound is  featured, the execution time  remains a problem. It ranges from few minutes to two days. This shows only the feasibility of the proposed method, not its practical application. To overcome this limitation, the completeness must be relaxed and some heuristic-based algorithms  have to be defined. Beam-search or Monte-Carlo Tree Search-based algorithms are good alternatives to the one we propose.}

\modif{Activation rule patterns are the simplest pattern language to deal with activation matrices since such patterns involve only conjunction of activated components.  Even simple, these activation rules are able to capture  the hidden features built by the GNN as witnessed by the experiments. We believe that more sophisticated pattern languages are possible for GNNs. For instance, we observed that taking into account the number of occurrences within a graph leads to better characterisations. This can be  integrated to the pattern language.  Considering the negation (i.e., the  absence of activations) is also promising and would offer a deeper description  of the internal mechanism of the GNNs. 
}

\modif{With \method, the activation rules are mined for each layer independently. As a consequence,  the relations between layers are not taken into account in the discovery of activation rules. This may lead to redundant results when considering all the layers. To avoid such redundancy, it is  necessary to take into account as prior knowledge the previous layers of a given layer.
}

\modif{Finally,  activation rules capture specific configurations in the embedding space of a given layer that is discriminant for the GNN decision. Experiments demonstrate that these rules  can be directly used to support instance-level model explanation. However,  activation rules cannot be easily interpreted by human beings because of the pattern language itself (i.e., conjunction of activated components of the hidden layers). The consideration of pattern languages with interpretable features makes it possible to characterize them. However, this second step can be improved by query the model itself. Indeed, the current characterization methods investigate a dataset generated from the support of the activation rules. The model should be considered in this step to have guarantee that the interpretable pattern that describes a rule well embeds in the subspace related to this rule. 
}

\bibliographystyle{spbasic}
\bibliography{bib}
\end{document}